%% file: main.tex
\documentclass{article}

\usepackage{PRIMEarxiv}

\usepackage[utf8]{inputenc} 
\usepackage[T1]{fontenc}    
\usepackage{hyperref}       
\usepackage{ragged2e}
\usepackage{microtype} 
\usepackage{url}            
\usepackage{booktabs}       
\usepackage{amsfonts}       
\usepackage{nicefrac}       
\usepackage{microtype}      
\usepackage{lipsum}
\usepackage{caption}
\usepackage{subcaption}
\usepackage{xcolor}
\usepackage[dvipsnames]{xcolor}
\usepackage{amsmath}
\usepackage{makecell}
\usepackage{fancyhdr}       
\usepackage{graphicx}       
\graphicspath{{media/}}     
\usepackage{placeins}
\usepackage{tablefootnote}
\usepackage{multirow}
\usepackage[absolute,overlay]{textpos}

\usepackage{wrapfig}

\usepackage{changepage}

\title{EngGPT2: Sovereign, Efficient and Open Intelligence
}
\author{
  Ciarfaglia G. \quad Rosanova A. \quad Cipolla S. \quad Bartoli J.\thanks{Equal contributions. Names are listed in alphabetical order.} \quad Di Domenico A.\footnotemark[1] \quad Fioroni C.\footnotemark[1] \quad Fontana A.\footnotemark[1]\\ \textbf{Scoleri M. R.\footnotemark[1]\quad Mone M. I. \quad Franchi D. \quad Del Gaudio M. C. \quad Leodori A. \quad Cinti F. \quad Capozzi M.} \\ \textbf{Baston C. \quad Picariello F. \quad Gabusi M. \quad Bonura S. \quad Morreale V. \quad Bailo I.}
  \\ 
  EngGPT Team @ Engineering Group\\
}

\begin{document}
\maketitle

\begin{textblock*}{9cm}(10.2cm,26.5cm) 
\raggedleft
\footnotesize
\textcolor{gray!85}{%
©Engineering Ingegneria Informatica S.p.A. All rights reserved.\\This document is provided for informational and research purposes only.%
}
\end{textblock*}

\vspace{-1.3cm} 


\vspace{12pt}

\begin{abstract}
\begin{adjustwidth}{-1.5cm}{-1.5cm}
\justifying
\sloppy
EngGPT2-16B-A3B\footnote[1]{\includegraphics[height=0.4cm]{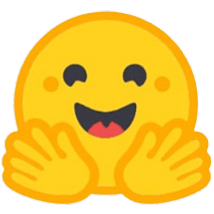}https://huggingface.co/engineering-group/EngGPT2-16B-A3B} is the latest iteration of Engineering Group’s Italian LLM and it’s built to be a Sovereign, Efficient and Open model. EngGPT2 is trained on 2.5 trillion tokens, less than Qwen3’s 36T or Llama3’s 15T, and delivers performance on key benchmarks, including MMLU-Pro, GSM8K, IFEval and HumanEval, comparable to dense models in the 8B–16B range, while requiring one-fifth to half of the inference power, and between one-tenth to one-sixth of the training data and consequent needed training power. Designed as a trained-from-scratch Mixture-of-Experts (MoE) architecture, EngGPT2 features 16 billion parameters with 3 billion active per inference, with expert sizes positioned between those used in GPT-OSS and Qwen3. Approximately 25\% of its training corpus consists of Italian-language data, to deliver strong capabilities for European and Italian NLP tasks among models of similar scale. This efficiency aims to position EngGPT2 as a key contributor to the growing portfolio of open-weight European models, combining performance and efficiency with full alignment to the EU AI Act. EngGPT2 is also a single model capable of multiple reasoning modes: non-reasoning, reasoning in Italian or English, and turbo-reasoning (a concise, bullet-point style reasoning available in both languages designed for real-time reasoning use cases). EngGPT2 aims to set a new standard for resource-conscious, high-performance LLMs tailored to European and Italian contexts.

\end{adjustwidth}
\end{abstract}

\begin{figure}[h]
    \centering
    \includegraphics[width=0.99\linewidth]{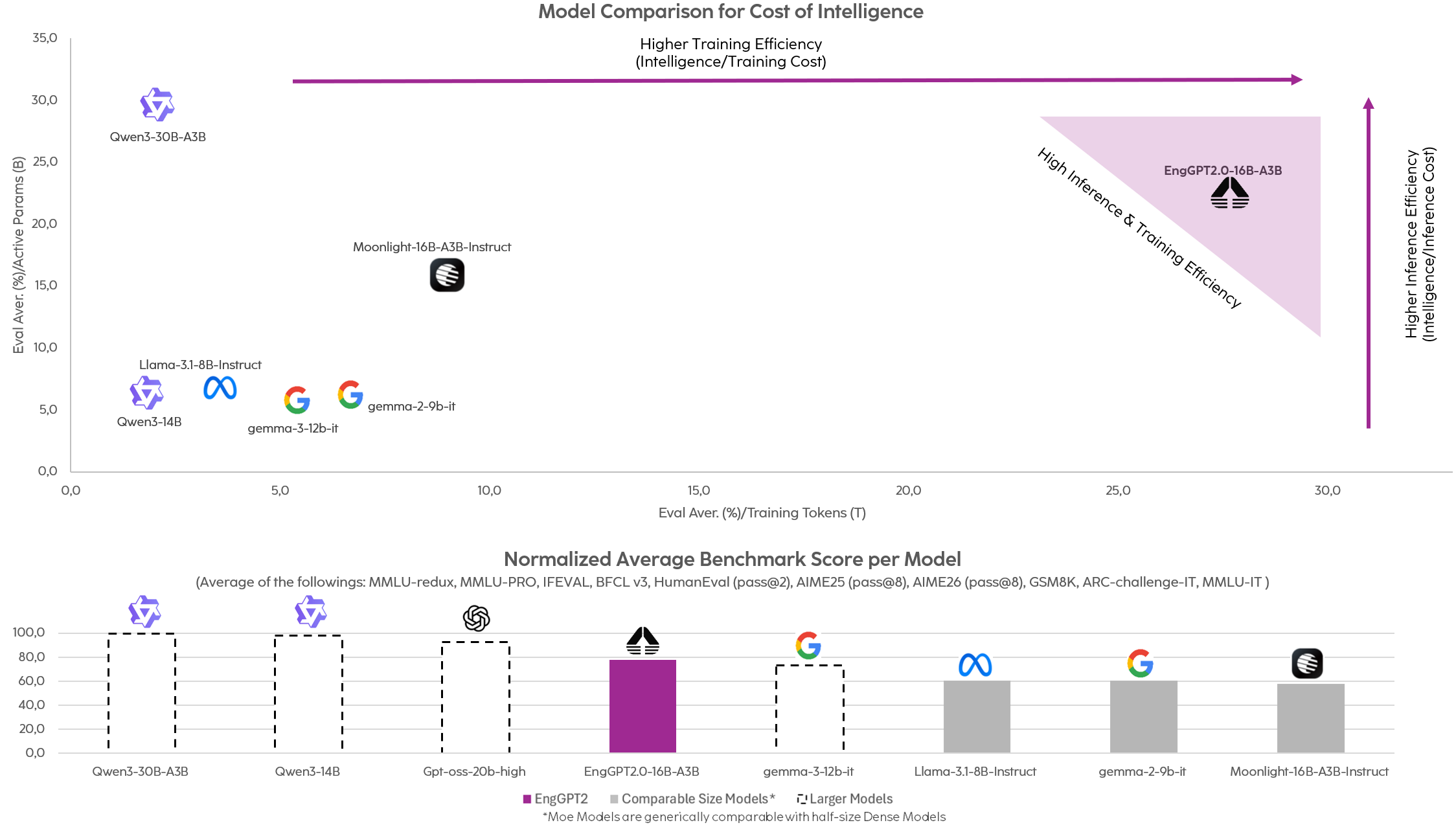}
    
    \caption{Comparison of model evaluation and efficiency. Top: model comparison under the Cost of Intelligence framework. Bottom: EngGPT2 average evaluation comparison against main open-weight comparable size models.}
    \label{fig:header_image}
\end{figure}


\input{Sections/01_introduction}
\input{Sections/02_architecture}
\input{Sections/03_eval_framework}
\input{Sections/04_pre_training}
\input{Sections/05_long_context}
\input{Sections/06_mid_training}
\input{Sections/07_post_training}
\input{Sections/08_evaluation}
\input{Sections/09_conclusion}
\input{Sections/10_Legal_notice}

\appendix

\include{Sections/11_appendixA}
\include{Sections/12_appendixB}

\include{Sections/13_appendixC}

\include{Sections/14_appendixD}
\include{Sections/15_appendixE}

\section{Acknowledgments} 
Part of this work is framed within the Project "AVANT" – Project no. IPCEI-CL\_0000005 - Application protocol no. 108421 of 14/05/2024 - CUP B89J24002920005 - Grant decree no. 1322 of August 8, 2024 - financed by the European Union – NextGenerationEU (IPCEI Funding).


\bibliographystyle{unsrt}  
\bibliography{references}

\end{document}

%% file: Sections/01_introduction.tex
\section{Introduction}
\subsection{Context and Scope}
Europe is steadily strengthening its capabilities in large language models, even if it still lags behind the major innovation hubs in the United States and China. In recent years, a small but growing set of national initiatives, driven by both industry and academia, has emerged to develop open, region‑focused foundation models. Although still limited in scale, these efforts mark an important starting point for building a European ecosystem committed to responsible, safe, and value‑aligned AI development.
As this ecosystem takes shape, generative AI is rapidly becoming a strategic asset for both public administrations and private enterprises. Cultivating strong local expertise is essential: technological sovereignty, sustained innovation, and seamless regulatory compliance all depend on Europe’s ability to design, operate, and govern AI systems within its own principles, standards, and legal frameworks.

Within this context, Engineering Group contributes not only as a technology integrator but as a true solutions builder, combining deep consulting expertise with engineering excellence to guide organizations in defining, designing, and deploying the AI solutions best suited to their specific needs. Our close collaboration with public institutions and industrial partners gives us direct insight into emerging requirements and real‑world operational constraints. This vantage point reinforces the importance of maintaining full control over the entire development pipeline, from initial design to deployment, and of investing in internal capabilities to build, adapt, and continuously refine customized AI systems rather than relying solely on off‑the‑shelf models.

In this spirit, we introduce EngGPT 2, our Large Language Model designed to strengthen Engineering’s technological stack and to contribute to the broader European effort to build a sovereign and open AI ecosystem.
Compliance with the EU AI Act is a foundational pillar of this sovereign and open approach. We adopt rigorous transparency practices, including comprehensive technical documentation and the publication of the model on the Hugging Face Hub, to support transparency, accountability, and regulatory compliance.
Beyond sovereignty and openness, EngGPT 2 strongly emphasizes efficiency, leveraging advanced architectural choices such as mixture‑of‑experts (MoE), dual‑reasoning flows, and native agentic integrations. These techniques enable significant performance gains, both in speed and in cost of operation, while supporting scalable, adaptable configurations tailored to enterprise and public‑sector environments.

Through this work, we aim to deliver not only advanced technological solutions but also a concrete contribution to a sovereign, efficient, and future‑ready European AI landscape.

\subsection{EngGPT 2 Overview}
We introduce EngGPT2‑16B‑A3B, a highly sparse, medium‑scale Mixture of Experts (MoE) language model designed as a flexible and efficient foundation for research and advanced downstream applications. Engineered to balance computational efficiency with strong general‑purpose reasoning, EngGPT2‑16B‑A3B supports a wide spectrum of capabilities, including structured reasoning, domain adaptation, tool‑augmented workflows, controlled generation, and general conversational proficiency. The EngGPT2‑16B‑A3B release provides full insight into its development pipeline, covering architectural decisions, data preparation, training dynamics, and evaluation methodology across the entire lifecycle of the model.

The project is centered on the principle that efficient parameter utilization through sparse expert-based architectures can deliver competitive performance while maintaining computational tractability \cite{shazeer2017outrageously}. Unlike dense models that activate all parameters during inference, the proposed MoE architecture enables selective activation of a subset of specialized sub-models (experts) on a per-token basis, thereby reducing computational overhead while maintaining expressive capacity. This approach aligns with recent industry advances demonstrating that carefully designed sparse models can achieve performance parity with significantly larger dense systems while incurring substantially lower training and inference costs \cite{Moe1, Moe2, Moe3,yang2025qwen3,agarwal2025gpt}.

The architecture is inspired by Qwen3 \cite{yang2025qwen3} and GPT-OSS \cite{agarwal2025gpt}. From Qwen3, we adopt hybrid attention and Grouped-Query Attention (GQA) for memory-efficient inference and sparse MoE routing enabling scalable parameters \cite{yang2025qwen3}. GPT-OSS influences the design with fewer, larger experts for an efficient MoE setup \cite{agarwal2025gpt}. The tokenizer draws from Mistral \cite{mistral_nemo}, chosen for its strong multilingual support, especially effective with Italian and other Romance and Germanic languages. 

The training process is structured around four distinct but interconnected phases, each with specific objectives and methodological approaches: 

\begin{enumerate}
\item \textbf{Pre-training: Language Foundation}

The model begins by learning fundamental language skills through self-supervised learning on large, diverse, and mostly raw text datasets. This phase emphasizes English and Italian to support multilingual generalization. Data from various public sources such as books, web corpora, scientific documents, and code are carefully curated to build a broad and diverse knowledge base. This foundational phase establishes the core language understanding that subsequent training phases cannot fully replicate. 

\item \textbf{Long-Context Adaptation: Extended Context Pretraining}

Before entering the mid-training stage, the model undergoes a dedicated phase aimed at adapting it to substantially extended context windows. This phase leverages targeted datasets with long documents to teach the model to maintain coherence, stability, and effective information retrieval over long sequences. The objective is to ensure reliable handling of contexts up to 32768 tokens and beyond.

\item \textbf{Mid-Training: Consolidation and Capability Enhancement}

This intermediate training stage focuses on consolidating knowledge and enhancing capabilities. Training centers on high-quality, carefully curated datasets that emphasize data quality over volume, targeting improvements in reasoning and linguistic precision. Logical, mathematical, and problem-solving skills are intensified by integrating specialized reasoning datasets, aiming to boost the model's capacity in these areas.

\item \textbf{Post-Training: Alignment and Instruction Following}

The final phase transitions the model into a chat-oriented, instruction-following system through supervised fine-tuning (SFT) and alignment techniques. Model merging is also applied to combine the strengths of SFT and alignment models. During this phase, the chat-template is formalized, and compatibility with function calling and the MCP server is ensured, enabling advanced agent-oriented interactive capabilities. 

\end{enumerate} 

The entire training pipeline was executed on an HPC infrastructure, scaling up to 128 nodes, each equipped with 4 NVIDIA A100 GPUs.
Our training pipeline leverages the Megatron framework \cite{shoeybi2019megatron}, a highly optimized and scalable system for training large transformer models across multiple GPUs via a combination of tensor, pipeline, and data parallelism. Building upon Megatron, we use custom code derived from the SmolLM3 project \cite{huggingface2025smollm3}, which incorporates modular enhancements and optimizations tailored for sparse Mixture of Experts (MoE) architectures. This hybrid framework allows efficient large-scale training while supporting advanced features such as long context windows, expert routing, and seamless integration with downstream fine-tuning and alignment workflows. 

On standard benchmarks, our model delivers performance that is fully comparable to dense baseline models of similar size. Beyond absolute benchmark scores, our model distinguishes itself even more clearly when metrics are normalized for training efficiency and inference cost. To capture these aspects, we define two composite metrics: one measuring capability per unit of training signal, and another measuring capability per active parameter at inference time. Under both normalizations, the model consistently outperforms larger dense baselines when compared at equivalent units of training or inference compute. These normalized perspectives highlight the core strength of our approach: delivering superior capability for a given compute budget, rather than merely achieving good absolute scores.

The technical report is structured to comprehensively document the project and includes the following chapters: 

\begin{itemize}
    \item \textbf{Architecture}: Describes the design principles and structural characteristics of the EngGPT 2 model, outlining the rationale behind its core architectural choices. 
    
    \item \textbf{Evaluation Framework}: Presents the general methodology used to assess the model across training phases, clarifying the evaluation principles and comparison criteria.
    
    \item \textbf{Pre-Training}: Summarizes how the foundational training of EngGPT 2 was conducted, outlining the multi-stage process, dataset mixture, and general training dynamics that shaped the base model. 
    
    \item \textbf{Long-Context Adaptation}: Explains the dedicated phase used to extend the model’s effective context window, describing how EngGPT 2 was adapted to handle significantly longer sequences. 
    
   \item \textbf{Mid-Training}: Describes the intermediate refinement stage focused on consolidating model behavior—particularly reasoning—under controlled conditions and preparing the model for subsequent alignment phases. 
    
    \item \textbf{Post-Training}: Outlines the steps performed after mid-training, including supervised fine-tuning, preference optimization, and the introduction of structured reasoning and interaction formats that enable final deployment capabilities.
    
    \item \textbf{Final Benchmarking and Comparative Analysis}: Reports how the finished model was evaluated against reference baselines.
    
    \item \textbf{Conclusion}: Summarizes the overall findings of the project and highlights the overarching direction for future model development.
    
    \item \textbf{Legal Notice}: Outlines the copyright terms, usage restrictions, disclaimers, and legal conditions governing the distribution and use of this document.
\end{itemize}

\subsection{Training Compute Assessment and GPAI Classification}
The full training pipeline required approximately 250,000 GPU hours, for an estimated total cost of €500,000, assuming an average cost of €2 per GPU hour. Pre-training represented the most computationally intensive stage, accounting for roughly 235,000 A100 GPU hours. This phase ran for approximately 23 days, scaling up to a maximum of 128 nodes (512 GPUs). The observed average Model FLOPs Utilization (MFU) was around 21\%, with peak values reaching 31\% during the most efficient segments of the run. The long-context training phase required approximately 2,000 GPU hours and was completed in about 20 hours, using up to 128 nodes (512 GPUs). Due to increased sequence lengths and reduced parallel efficiency, the average MFU during this phase was lower, at approximately 14\%. The mid-training phase consumed roughly 12,000 GPU hours over a period of 7 days, operating on a reduced configuration of 16 nodes (64 GPUs). The average MFU observed in this stage was approximately 8\%, reflecting additional computational overhead introduced by data-dependent processing and masking strategies, which reduce peak hardware utilization in favor of improved training dynamics. Finally, the post-training stages, including fine-tuning and post-training alignment, accounted for approximately 4,000 GPU hours. These activities spanned around 10 days and were executed using variable configurations ranging from 4 to 16 nodes. The MFU during post-training is highly variable across different phases but consistently remains well below 10\%. This is expected, as post-training workloads are not compute-bound: they involve autoregressive generation, frequent synchronization, evaluation steps, and small batch sizes, all of which naturally lead to low hardware utilization. 

To estimate the computational footprint of our training pipeline, we assumed a peak performance of 312 TFLOPS per NVIDIA A100 GPU (FP16). For each phase of training, we computed an average effective throughput by multiplying this peak value by the empirically observed GPU utilization rate. Specifically, we used 65.52 TFLOPS (21\% of peak) for the pre‑training phase, 43.68 TFLOPS (14\%) for long‑context training, 25 TFLOPS (8\%) for mid‑context training. For each phase, the total computational cost was obtained by multiplying the corresponding average TFLOPS by the total number of GPU‑hours, converted into seconds (i.e., GPU‑hours × 3600). This yields a total of $5.5\times{10}^{10}$ TFLOPs for pre‑training, $3.1\times10^{8}$ TFLOPs for long‑context training, $1.1\times10^{9}$ TFLOPs for mid‑context training.
The post‑training stage consisted of multiple sub‑phases with different measured MFU values; therefore, we directly report the aggregated computational cost across the 4000 GPU‑hours, which amounts to $3.6\times{10}^{8}$ TFLOPs in total. Summing across all training phases, the total cumulative compute amounts to $5.7\times{10}^{22}$ FLOPs.

Our model qualifies as a GPAI model under the European AI Act framework, as outlined in \cite{AI_ACT}. However, it is not subject to the obligations related to system risk reporting, since its cumulative training compute of $5.7\times{10}^{22}$ FLOPs is well below the ${10}^{25}$ FLOPs threshold specified in the regulation.
\clearpage

%% file: Sections/02_architecture.tex
\section{Architecture}
Our model leverages a Mixture-of-Experts (MoE) transformer architecture with 24 layers. Each layer contains 64 experts, with 8 experts activated per token through dynamic routing. No parameters are shared across experts, which promotes greater specialization. The attention system uses Grouped Query Attention (GQA) \cite{ainslie2023gqa}, and each layer incorporates SwiGLU activations \cite{dauphin2017}, Rotary Positional Embeddings (RoPE) \cite{su2023re}, and RMSNorm \cite{jiang2023} with pre-normalization to ensure training stability and effective representation. 

\begin{wrapfigure}{r}{0.6\textwidth}
    \centering
    \includegraphics[width=1\linewidth]{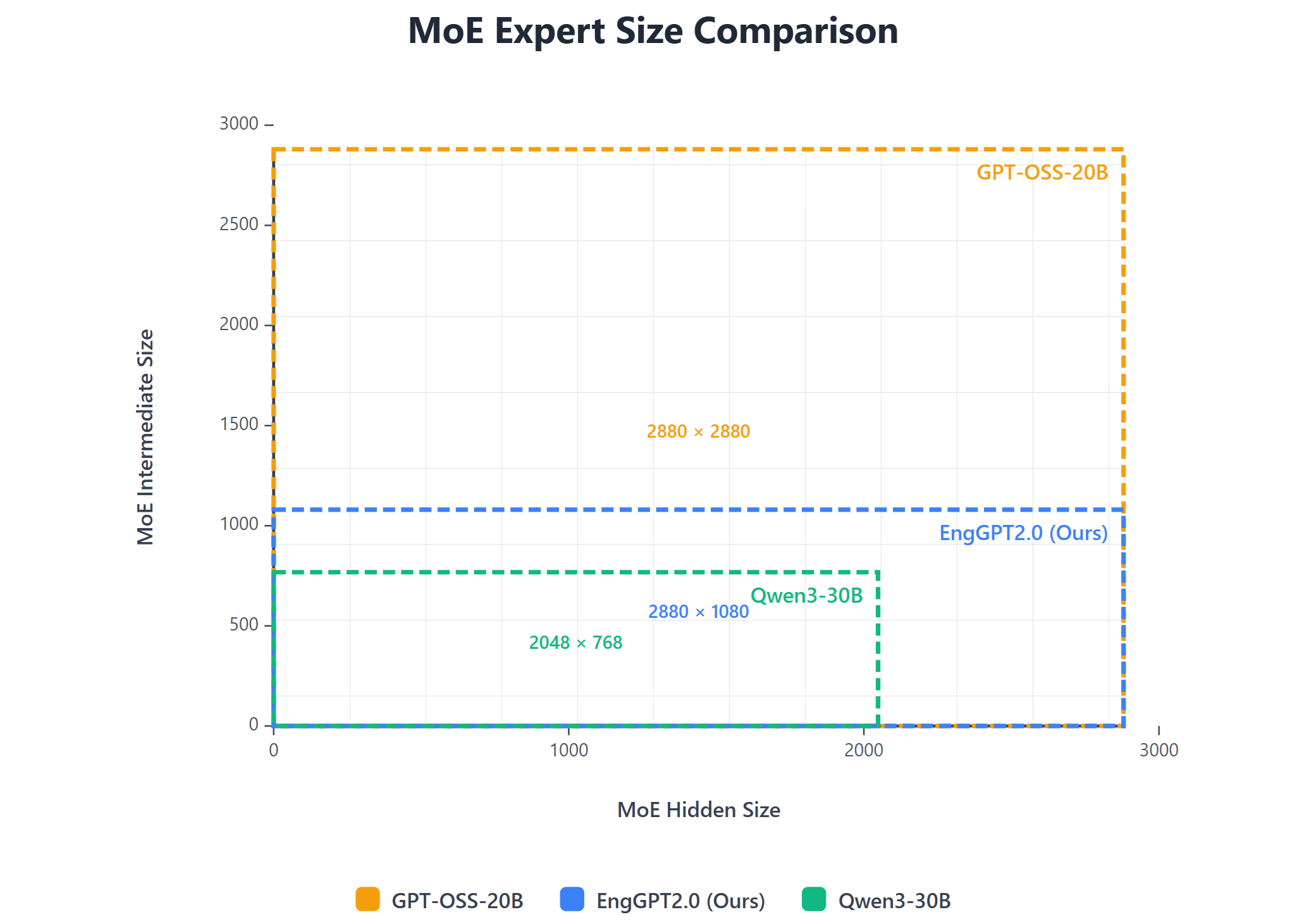}
    \caption{Model expert size comparison}
    \label{fig:expert_size_comparison}
\end{wrapfigure}

In our design, the individual experts are configured with hidden size $H$ and MoE intermediate size $m$, chosen to lie between Qwen3-30B-A3B \cite{yang2025qwen3} and Openai-gpt-oss-20b \cite{agarwal2025gpt}. This decision is driven by the need to maintain a minimum of approximately 3B active parameters per forward pass, ensuring sufficient model capacity for high-quality reasoning, while keeping the total model size manageable within our training budget constraints. By selecting intermediate-sized experts, we strike a balance between efficiency and expressiveness, allowing the model to perform effectively without incurring excessive computational or memory costs. Finally, we sought to identify a Mixture-of-Experts configuration that is original and uniquely tailored to our model. In \autoref{fig:expert_size_comparison}, we illustrate a visual comparison of different model expert sizes. See \autoref{tab:architectural_parameters} and \autoref{tab:moe_comparison} for further details about the model's architecture.


The model’s tokenizer employs a vocabulary size that is expanded to 131084 tokens. Beyond the standard vocabulary, it incorporates 12 additional specialized tokens specifically designed to support reasoning and tool-calling capabilities. 

\begin{table}[h]
\centering
\setlength{\arrayrulewidth}{0.4pt} 
\setlength{\tabcolsep}{15pt} 
\renewcommand{\arraystretch}{2} 

\begin{tabular}{c|c|c|c}
\textbf{Feature} & \textbf{Value} & \textbf{Feature} & \textbf{Value} \\
\specialrule{1.5pt}{0pt}{0pt} 
Number of Layers & $24$ & Num attention heads  & $32$ \\
Total number of experts & $64$ & Num key value heads & $4$ \\
Experts activated/token & 8 & Context length & $32768$ \\
Vocabulary size & $131084$ & & \\
\end{tabular}

\vspace{5mm}
\caption{Architectural Parameters}
\label{tab:architectural_parameters}
\end{table}

\begin{table}[h]
\centering
\resizebox{\textwidth}{!}{
\setlength{\arrayrulewidth}{0.4pt}
\setlength{\tabcolsep}{5pt}
\renewcommand{\arraystretch}{2}

\begin{tabular}{c|c|c|c|c|c|c|c|c}
\textbf{Model} & \shortstack{\textbf{\# Params} \\ (Total / Active)} & 
\shortstack{\textbf{Single Expert} \\ \textbf{Size}} &
\textbf{H/m} &
\shortstack{\textbf{\% Active} \\ \textbf{Params}} &
\textbf{Layers} &
\shortstack{\textbf{Experts} \\ (Total / Activated)} &
\shortstack{\textbf{Heads} \\ (Q /KV)} &
\textbf{Vocab size} \\
\specialrule{1.5pt}{0pt}{0pt} 
\textbf{EngGPT2-16B-A3B} & 16B / 3B & 9.3M & 2880 / 1080 & $20.27\%$ & 24 & 64 / 8 & 
32 / 4 & 131084 \\
\textbf{Qwen3-30B-A3B} & 30B / 3B & 4.7M & 2048 / 768 & $10.98\%$ & 48 & 128 / 8 & 
32 / 4 & 151936 \\
\textbf{Openai-gpt-oss-20B} & 21B / 3.6B & 24.9M & 2880 / 2880 & $17.30\%$ &
24 & 32 / 4 & 64 / 8 & 201088 \\
\end{tabular}
}
\vspace{5mm}
\caption{MoE Comparison}
\label{tab:moe_comparison}
\end{table}



%% file: Sections/03_eval_framework.tex
\section{Evaluation Framework}
\label{sec:eval_framework}
Our evaluation methodology provides a comprehensive, phase-aware assessment of the model’s capabilities, combining standardized benchmark measurements with targeted analyses tailored to the objectives of each stage of the training pipeline. The approach prioritizes reproducibility, comparability with prior work, and a clear separation between capabilities acquired during base pre-training and behaviors introduced through subsequent alignment and refinement stages.

Quantitative evaluation is primarily conducted using the lm-evaluation-harness framework \cite{sutawika2025eleutherai}, ensuring consistent and robust measurement across a broad range of academic and open-source language model benchmarks. Function-calling performance is assessed through the BFCL V3 suite using the EvalScope framework \cite{evalscope}. All experiments are executed under controlled and fully documented serving configurations to guarantee transparency and replicability.

The evaluation framework spans the entire training lifecycle, with a level of depth proportional to the model’s maturity at each phase. During early pre-training checkpoints and intermediate refinement stages, the assessment is intentionally limited to a core subset of key performance indicators (KPIs). These metrics are selected to validate training stability, verify correct optimization dynamics, and monitor the progressive emergence of target capabilities. At this stage, the objective is not comparative benchmarking, but controlled validation of training quality and trajectory.

A complete and systematic benchmarking analysis is presented in \autoref{sec:benchmarking}. This section provides an extensive comparative evaluation that integrates standardized benchmark suites with qualitative and capability-oriented assessments. The analysis covers general knowledge, instruction following and alignment, code generation, mathematical and logical reasoning, Italian-language performance, and function-calling capabilities. It also includes direct comparisons with architectures of comparable parameter scale as well as with larger baseline models, offering both an assessment of competitiveness within the same model class and an evaluation of efficiency and performance trade-offs relative to more computationally demanding systems.

\subsection{Evaluation Approach}
During pre-training, evaluation focuses on tracking learning progression and convergence dynamics through foundational benchmarks that measure intrinsic language modeling quality, general knowledge acquisition and long-context understanding. These assessments are primarily intended to monitor capability growth and convergence dynamics rather than to provide competitive comparisons. 

Post-training evaluation, by contrast, is oriented toward final-model assessment and external benchmarking. At this stage, the model is evaluated on a broader and more usage-oriented suite of benchmarks covering knowledge, structured reasoning, instruction following, code generation, multilingual performance, and tool-calling ability. These results are used both to measure downstream readiness and to enable direct comparison against peer models. Multilingual evaluation—particularly for Italian—is treated as an integral component of the final assessment, ensuring that deployment-oriented performance is validated beyond English-only benchmarks. 

When benchmarking against external models, we systematically re-evaluated all systems under two distinct serving configurations to ensure strict and transparent comparability. First, all models were evaluated using a fully standardized generation setup, with identical sampling hyperparameters (temperature, top-p, top-k, and min-p) applied across all systems. This unified configuration eliminates variability introduced by decoding strategies and ensures that performance differences reflect intrinsic model capability rather than sampling artifacts. Second, whenever officially recommended or best-performing serving configurations were provided in the respective model documentation, we additionally report results obtained under those optimal settings. This dual evaluation protocol—standardized configuration for controlled comparison and optimal configuration for capability estimation—ensures both methodological fairness and practical relevance, reducing evaluation bias while preserving fidelity to each model’s intended deployment regime. 

To better contextualize raw benchmark results with respect to efficiency, we introduce two additional composite metrics that are of particular interest to our use case. The first metric captures a normalized average performance score, computed as the mean of the selected KPIs after normalization, divided by the total number of training tokens. This metric is intended to approximate performance per unit of training signal, highlighting how effectively the model converts training data into downstream capability.

The second metric follows a similar aggregation strategy but normalizes the averaged KPI score by the number of active parameters at inference time, rather than by total model size. This metric is especially relevant for MoE architectures, as it directly reflects the effective inference footprint and provides a more faithful measure of performance-to-compute efficiency than raw parameter counts. 

Taken together, these metrics allow us to move beyond absolute benchmark scores and explicitly capture the trade-off between capability and cost. Under both normalizations, our model consistently achieves performance levels that are close to those of larger dense baselines, while requiring significantly fewer training tokens and substantially fewer active parameters during inference. This results in markedly lower training costs and reduced inference latency and compute consumption, without a proportional degradation in accuracy or reasoning quality.

\subsection{Benchmark suite}
We evaluate core linguistic, reasoning, and task-oriented capabilities through a diverse suite of benchmarks covering general knowledge, instruction following, code generation, mathematical reasoning, multilingual performance, and function-calling ability.

\begin{itemize}
    \item \textbf{General Knowledge:} We assess broad factual knowledge and domain-specific expertise using MMLU \cite{hendrycks2020measuring}, MMLU-Pro \cite{wang2024mmlu}, and MMLU-Redux \cite{gema2025we}. These datasets span a wide range of academic disciplines and difficulty levels, enabling a holistic evaluation of factual recall, professional-level knowledge, and robustness across subject domains.
    \item \textbf{Instruction Following / Alignment:} Instruction-following behavior and alignment with user intent are evaluated using IFEval \cite{zhou2023instruction}. We report loose prompt accuracy, which measures the model’s ability to correctly interpret instructions, adhere to structural constraints, and produce outputs aligned with formatting and semantic requirements. This benchmark is particularly relevant for real-world deployment scenarios involving structured prompts and constrained outputs.
    \item \textbf{Code Generation:} Programming and code synthesis capabilities are evaluated using HumanEval \cite{chen2021evaluating}. We report pass@2, which estimates the probability that at least one of two generated samples passes all unit tests. This metric reflects realistic code-generation workflows where multiple candidate solutions may be sampled.
    \item \textbf{Reasoning:} For evaluating mathematical and logical reasoning skills, we employ high-level math benchmarks including AIME25 \cite{AIME25}, AIME26 \cite{AIME26}, GSM8K \cite{cobbe2021training}. Each AIME year includes Part I and Part II, for a total of 30 problems. For each problem, we compute pass@8, providing a robust estimate of reasoning reliability under stochastic decoding.
    \item \textbf{Italian Language:} To assess multilingual capabilities—particularly for deployment in the Italian market—we include Italian-adapted benchmarks: ARC-Challenge-IT \cite{m_arc_it} and MMLU-IT \cite{mmlu_it}. These benchmarks evaluate knowledge recall across a broad range of subjects, problem-solving and analytical reasoning expressed in Italian, and robustness in handling domain-specific terminology and culturally grounded content. This ensures that performance is not disproportionately optimized for English and that quality, precision, and instruction adherence are preserved in Italian-language deployment scenarios.
    \item \textbf{Function Calling:} The ability to correctly invoke external tools and APIs is evaluated using the Berkeley Function-Calling Leaderboard (BFCL v3) \cite{bfcl_v3},  assessed via the EvalScope framework \cite{evalscope}. The benchmark covers AST-based evaluation (both non-live and live), relevance detection, and multi-turn interaction scenarios, providing a comprehensive assessment of the model's ability to interface with structured APIs in realistic deployment settings.
\end{itemize} 

\subsection{Evaluation Protocol for Final Benchmarking}
All post-training evaluations are conducted exclusively in generative mode, where the model produces free-form outputs until EOS token is generated or a maximum length limit is reached (35000 tokens or the model’s maximum context window if lower for reasoning models). This setup differs from the multiple-choice evaluation used during pretraining and better reflects real world downstream usage including open-ended reasoning, instruction following, structured and long-form generation. Evaluations are performed in the appropriate zero-shot, ensuring methodological adherence and comparability across models. For each model–benchmark pair, we verified that evaluation metrics accurately captured the intended answers by aligning extraction rules and stop sequences with the task format, validating answer normalization procedures, and checking for truncation or misparsing.


To ensure strict cross-model comparability, we evaluate all models using the same generation hyperparameters adopted for EngGPT2-16B-A3B: temperature = 0.6, top-p = 0.95, top-k = 20, and min-p = 0. 
In addition, to ensure that the reported results reflect each model’s maximum achievable capability under realistic serving conditions, we also evaluate every model in its best-performing configuration, as detailed in \autoref{app:eval_config}. 

This dual reporting strategy (optimal configuration vs. standardized configuration) provides both a fair estimate of maximum practical capability and a controlled comparison under identical sampling conditions.

We aim to benchmark EngGPT2‑16B‑A3B against models of comparable size. Within this category, the closest MoE reference is Moonlight‑16B‑A3B‑Instruct \cite{moonlight}, which matches both the architectural class and the overall parameter scale. Beyond Moonlight, however, few MoE models exist in this size range, limiting direct comparisons within the MoE family.

Given the scarcity of similarly sized MoE baselines, we adopt a heterogeneous selection strategy that incorporates dense models of approximately 8B parameters as our primary “comparable‑sized” baselines. This approach is supported by evidence—most notably \cite{yang2025qwen3}—showing that MoE architectures often achieve performance comparable to dense models with roughly half as many parameters. In that study, for example, Qwen3‑30B‑A3B delivers results close to the dense Qwen3‑14B, reinforcing the validity of comparing EngGPT2‑16B‑A3B with dense models in the ~8B range. Following this rationale, our comparable‑sized baselines include Llama‑3.1‑8B‑Instruct \cite{grattafiori2024llama} and gemma‑2‑9b‑it \cite{team2024gemma}.

To contextualize performance beyond this range, we further include a set of larger dense and MoE models. Here, our selection is again deliberate: among the many available options, we specifically choose Qwen3‑14B and Qwen3‑30B‑A3B, as their relationship has been well‑studied and provides a clear and meaningful MoE‑vs‑dense comparison point, consistent with the observations from \cite{yang2025qwen3}. Alongside these, we include gemma‑3‑12b‑it \cite{gemma_2025} and GPT‑OSS‑20B\cite{agarwal2025gpt}. This broader set enables us to characterize how EngGPT2‑16B‑A3B scales relative to both larger dense architectures and substantially larger MoE systems, while maintaining conceptual coherence across the chosen baselines.

%% file: Sections/04_pre_training.tex
\clearpage
\section{Pre-Training}
The pre-training stage forms the cornerstone of EngGPT’s linguistic and multilingual foundation. It was structured as a three-phase process, each serving distinct objectives and executed under strict computational budget constraints, limiting total training to approximately 2.5 trillion tokens. The first phase also served as a warmup and baseline step, providing an initial model for validating the entire mid- and post-training pipelines.
\begin{itemize}
\item Phase 1: Warmup and Initial Model (600B tokens): Establishes a stable baseline checkpoint and verifies the reliability of the end-to-end training infrastructure.
\item Phase 2: Main Scale-up (1.5T tokens): Expands corpus coverage and consolidates the model’s core linguistic capabilities.
\item Phase 3: High-Quality Refinement (400B tokens): Focuses on curated, high-quality datasets to strengthen reasoning, coding, and domain precision before transitioning to mid-training.
\end{itemize}

This phased strategy enabled efficient utilization of available GPU-hour allocations while ensuring a robust baseline for downstream validation and development.

\subsection{Pre-Training Data}
The pre-training corpus was constructed by combining several large-scale, publicly available datasets covering web text, educational material, mathematical content, code, PDF documents, and synthetic data. Specifically, we leveraged FineWeb-2, FineWeb-Edu, FineMath, FinePDFs, StarcoderData, and Nemotron-Pretraining-SFT-v1. The composition of the data mixture across the different training stages is reported in \autoref{tab:pre_train_datasets}.

Each dataset was used in accordance with its respective license terms. The training corpus includes a combination of datasets released under open licenses and/or datasets governed by usage-restricted agreements, as well as curated collections of permissively licensed or open-access content. Where datasets included heterogeneous sources (e.g., code or web data collections), additional filtering and selection procedures were applied to retain only content compliant with the applicable licensing terms. A detailed description of each dataset, including data sources, scale, filtering strategies, and privacy considerations, is provided in Appendix \ref{appendix:datasets}.

All pre-training data underwent standardized preparation and tokenization pipelines executed directly on the HPC infrastructure.   Specifically, tokenization and preprocessing were performed leveraging Megatron’s native data utilities combined with the Datatrove  library for efficient sharded data handling.

The tokenizer is inspired from Mistral\cite{mistral_nemo}, chosen for its proven multilingual performance and strong encoding efficiency across Italian, English, French, Spanish, and German. This tokenizer supports substantial vocabulary overlap across Romance and Germanic languages, reducing fragmentation and improving cross-lingual generalization.

Data ingestion, shuffling, and packing workflows were orchestrated to ensure uniform token distribution and consistent throughput across hundreds of data-parallel workers. The result was a reproducible and scalable pre-training pipeline capable of handling multi-terabyte tokenized datasets while maintaining data determinism required for checkpoint reproducibility.

Rather than designing an ad-hoc data cleaning pipeline, we relied on the data curation, filtering, deduplication, and PII mitigation procedures provided by the original dataset creators, which include language identification, quality scoring, near-duplicate removal, decontamination against standard evaluation benchmarks, and license compliance checks. This choice allows us to build upon well-established and transparent preprocessing pipelines while ensuring reproducibility and legal clarity.

In addition to standard data cleaning procedures, we implemented a dedicated copyright-filtering pipeline to mitigate the inclusion of protected material. This pipeline combines rule-based heuristics, pattern matching to assign each record a composite copyright risk score. Records exceeding a predefined risk threshold are automatically excluded from the final corpus. By integrating multiple signals into a unified scoring framework, this approach enables systematic identification and removal of documents likely to contain copyrighted content, thereby reducing potential infringement and improving the overall compliance of the dataset. This procedure is further explained in Appendix \ref{appendix:pretrain_filtering}.

\begin{table}
\centering
\setlength{\arrayrulewidth}{0.4pt} 
\setlength{\tabcolsep}{15pt} 
\renewcommand{\arraystretch}{2} 

\begin{tabular}{c|c|c|c|c}
\textbf{Dataset} & \shortstack{\textbf{License}} & \shortstack{\textbf{Stage 1} \\ \textbf{(600B token)}} & \shortstack{\textbf{Stage 2} \\ \textbf{(1.5T token)}} & \shortstack{\textbf{Stage 3} \\ \textbf{(400B token)}} \\
\specialrule{1.5pt}{0pt}{0pt} 
\shortstack{\textbf{Fineweb-edu} \textbf{(english)}} & odc-by & 52\% & 48\% & 21\% \\
\textbf{Fineweb2 - ita} & odc-by & 23\% & 12\% & 15\% \\
\textbf{Finepdf - ita} & odc-by & 0\% & 5\% & 5\% \\
\textbf{Fineweb2 - rom lang} & odc-by & 12\% & 12\% & 6\% \\
\textbf{Finemath} & odc-by & 1\% & 3\% & 0\% \\
\shortstack{\textbf{Nemotron-Pretraining} \\ \textbf{(SFT-v1 - Math)}}  & NVIDIA\tablefootnote{\label{NVIDIA_data}Data governed by the NVIDIA Data Agreement for Model Training \\(https://huggingface.co/datasets/nvidia/Nemotron-Pretraining-Dataset-sample/raw/main/LICENSE.md)} & 0\% & 4\% & 17\% \\
\textbf{StarCoder} & Misc\tablefootnote{\label{misc_data}The dataset was filtered to retain only content from repositories identified as open-access / permissively licensed, in accordance with the dataset’s documentation and licensing guidelines (https://huggingface.co/datasets/bigcode/the-stack)} & 12\% & 7\% & 12\% \\
\shortstack{\textbf{Nemotron-Pretraining} \\ \textbf{(SFT-v1 - Code)}} & NVIDIA\textsuperscript{\ref{NVIDIA_data}} & 0\% & 4\% & 10\% \\
\shortstack{\textbf{Nemotron-Pretraining} \\ \textbf{(SFT-v1 - General)}} & NVIDIA\textsuperscript{\ref{NVIDIA_data}} & 0\% & 5\% & 14\% 
\end{tabular}

\vspace{5mm}
\caption{Pretrain datasets. The listed datasets are subject to different licensing regimes, including open license, usage-restricted agreements and curated collections of permissively licensed content.}
\label{tab:pre_train_datasets}
\end{table}




\subsection{Pre-Training Stages}
The entire pre-training process was carried out on an HPC infrastructure, with each node equipped with four NVIDIA A100 GPUs, using the Megatron framework to support large‑scale distributed transformer training \cite{megatronLM}. The system employed a hybrid parallelization strategy combining tensor, data, pipeline, and expert parallelism, with a pipeline parallelism degree of 2 and an expert parallelism degree of 4. This configuration ensured that expert layers remained primarily contained within the nodes of four GPUs, reducing inter‑node communication overhead and stabilizing overall throughput. Parallel folding was enabled across all stages \cite{MOEparallelFolding}, contributing to a sustained peak Model FLOPs Utilization (MFU) of approximately 31\%, which represented a well‑balanced equilibrium between computational efficiency and training stability for this hardware configuration.

All pre‑training stages were executed as independent training runs, each reinitialized with specific optimization schedules, parallel configurations, and expert‑routing parameters tailored to the progression of the model’s learning dynamics. Across all stages, training was performed on sequences of 4096 tokens.

\subsubsection{Stage 1 - Warmup and Initial Model}
The first stage served as both a system warmup and the creation of an initial stable checkpoint. It was executed on 64 nodes (256 NVIDIA A100 GPUs) and consisted of 75,776 optimization steps, targeting approximately 600 billion training tokens.

This phase adopted a warmup‑and‑decay learning‑rate schedule, beginning at $1.89 \times 10^{-7}$ and increasing linearly to $1.2 \times 10^{-4}$, after which the learning rate gradually decayed over the remainder of the training. During the initial 19,076 steps, the global batch size (GBS) was set to 1,024, supporting a smooth warmup of both model dynamics and distributed infrastructure. After the warmup, the GBS was increased to 2,048. Gradient accumulation began at 8 and was later increased to 16 to handle the higher effective batch size without exceeding memory limits. 

From a performance perspective, Stage 1 achieved an average throughput of 70.74 TFLOPs/GPU, corresponding to an MFU of 22.6\%. The effective training time—excluding overheads—was approximately 8 days.

Beyond throughput metrics, this stage validated the stability of large‑scale distributed training, ensured proper expert‑routing behavior, and produced the foundational checkpoint used in subsequent pre‑training phases. The load balancing loss coefficient for this phase was set to $1 \times 10^{-2}$, promoting uniform expert utilization during the early stages of learning.

\subsubsection{Stage 2 - Main Scale-up}
The second stage marked the main scale‑out effort of pre‑training, shifting focus toward broad multilingual generalization and long‑horizon optimization stability. Training was reinitialized on an expanded configuration of 128 nodes (512 GPUs), for a total of 182,212 optimization steps, processing approximately 1.5 trillion tokens.

In contrast to Stage 1, this phase employed a constant learning rate of $1 \times 10^{-4}$, maintaining consistent gradient dynamics over a prolonged training window. The global batch size was fixed at 2,048, and gradient accumulation remained at 8 across the entire run. To improve expert uniformity in the larger‑scale setup, the load balancing loss coefficient was reduced to $5 \times 10^{-3}$, which resulted in smoother routing behavior and reduced expert‑specific variance.

Average throughput during Stage 2 reached 63 TFLOPs/GPU (MFU 20.2\%), aligning with expectations for a model of this size operating on the increased GPU count. The stage delivered the bulk of the model’s representational capacity, forming the core of pre‑training before refinement.

\subsubsection{Stage 3 - High-Quality Refinement}
The third stage focused on refinement, domain specialization, and preparation for mid‑training. Like Stage 2, it was executed on 128 nodes (512 GPUs), comprising 63,578 optimization steps and processing approximately 400 billion tokens.

This phase adopted a linear learning‑rate decay schedule, starting from $1 \times 10^{-4}$ and decreasing to $4.52 \times 10^{-10}$, enabling increasingly stable and fine‑grained updates as training approached its endpoint. The global batch size was again set to 2,048, with gradient accumulation maintained at 8. The load balancing loss coefficient was further reduced to $1 \times 10^{-3}$, encouraging more adaptive expert specialization while mitigating routing imbalance during the final training phase. Average throughput during Stage 3 was 65.9 TFLOPs/GPU, corresponding to an MFU of 21.1\%.

This final pre‑training stage emphasized curated, high‑quality sources—particularly code, mathematics, and structurally rich language datasets—to enhance reasoning accuracy, symbolic manipulation, and structured task performance before transitioning into mid‑training.

\subsubsection{Throughput Dynamics across Stages}
\begin{figure}[h]
    \centering
    \includegraphics[width=0.72\linewidth]{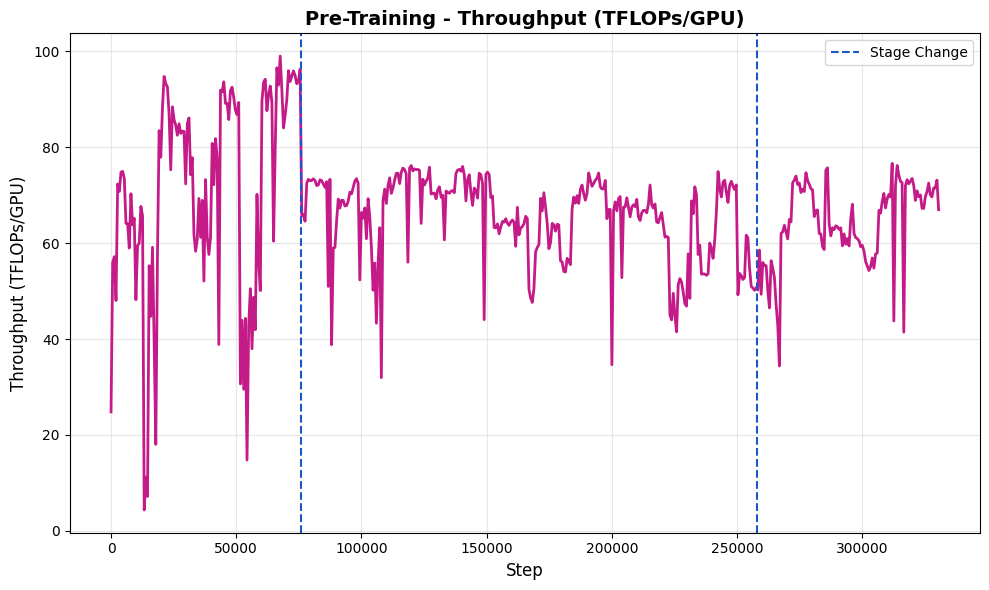}
    \caption{Throughput during the first three stages of pre-training.}
    \label{fig:pretrain_throughput}
\end{figure}

The throughput profile across the three pre‑training stages (see \autoref{fig:pretrain_throughput}) reflects the combined effects of gradient‑accumulation settings, GPU scaling, and learning‑rate schedules. The initial portion of the curve — corresponding to Stage 1 — displays a distinctive two‑regime behavior. During the earliest steps, throughput is noticeably lower and more volatile, primarily due to the use of a gradient accumulation factor of 8, which limits the effective batch size and results in more frequent synchronization across the 256‑GPU configuration. In this period, the training system is also still settling in: expert routing, activation patterns, and pipeline stage utilization are stabilizing, all of which contribute to the marked fluctuations visible in the early segment of the graph.

As Stage1 progresses and the gradient accumulation is increased to 16, throughput improves significantly. This transition is clearly visible in the chart, where throughput rises into a more stable and higher‑performance band. The larger effective batch size reduces communication overhead relative to compute time, allowing GPUs to operate with improved efficiency. Once this higher accumulation setting is in place, the remainder of Stage1 shows a more consistent throughput profile.

The first vertical dashed line marks the transition to Stage2. Here, the throughput stabilizes into a sustained plateau, reflecting the move to 512 GPUs and the use of a fixed gradient accumulation of 8 throughout the entire stage. Despite the lower accumulation factor relative to the latter part of Stage1, the throughput does not degrade significantly; the larger number of GPUs and the uniform training configuration help maintain a consistent efficiency level. The constant learning rate used in Stage2 further contributes to the steady behavior observed, minimizing optimizer‑induced variability.

The second dashed line marks the start of Stage3. As expected, the throughput pattern remains very similar to Stage2, since both stages use the same number of nodes (512 GPUs), identical parallelism configuration, and the same gradient accumulation factor. The only notable difference stems from the linear decay of the learning rate, which introduces mild fluctuations in the early portion of Stage3 but does not fundamentally alter the throughput regime. Once the decay stabilizes, throughput returns to the familiar band established during Stage2.

\subsection{Pre-Training Evaluation}
\begin{figure}[h]
    \centering
    \includegraphics[width=0.72\linewidth]{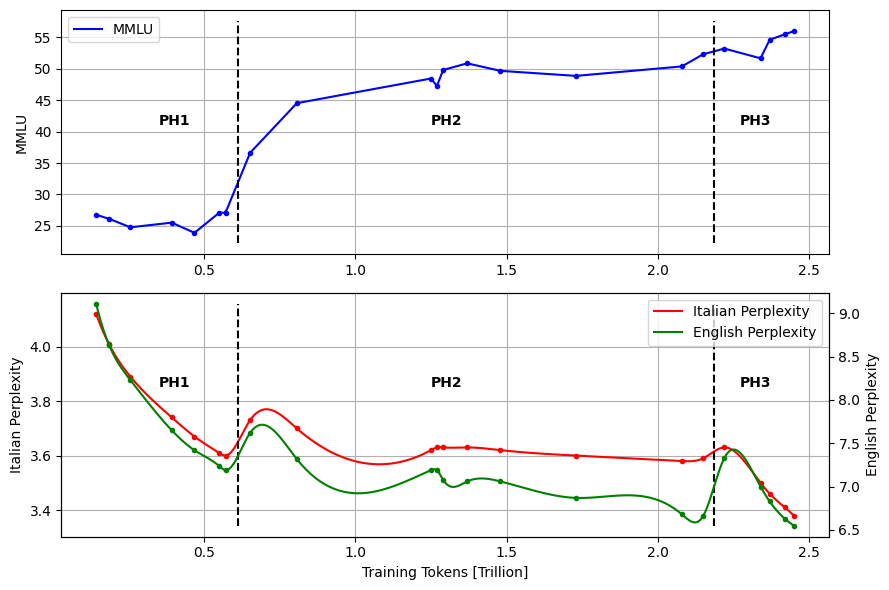}
    \caption{MMLU accuracy and perplexity across the first three pre-training phases as a function of the number of processed training tokens. MMLU evaluation is conducted on a fixed subset of the benchmark to monitor the emergence of general knowledge throughout training. English perplexity is measured on a fixed subset of WikiText-2, while Italian perplexity is computed on a fixed subset of a curated Wikimedia-based Italian corpus, following the same evaluation protocol used for English. The decreasing perplexity trend reflects progressively improved predictive performance on held-out data.}
    \label{fig:mmlu_pretrain}
\end{figure}

During the initial stages of pre-training, the model’s progression was continuously monitored through a combination of capability-oriented and intrinsic evaluation signals. In particular, across the first three pre-training phases, we tracked performance on MMLU as an external proxy for the emergence of general knowledge. MMLU was selected for this purpose, as it provides a task-level assessment of knowledge integration and multi-domain reasoning that is not directly captured by intrinsic language modeling metrics such as perplexity. As shown in \autoref{fig:mmlu_pretrain}, MMLU accuracy at early checkpoints was close to random-choice performance (approximately 25\%), reflecting the limited task-specific competence of the model at this stage. With continued training, MMLU performance increased steadily, reaching approximately 55\% by the end of the third pre-training phase, indicating substantial gains in general knowledge acquisition and reasoning potential as the model was exposed to increasing amounts of training data.

In parallel, we monitored language modeling quality through perplexity measurements, following the standard evaluation procedure described in the Hugging Face Transformers documentation \cite{hf_perplexity_doc}. 
For the English language, perplexity was evaluated on the WikiText-2 dataset \cite{wikitext2_dataset}, while Italian perplexity was measured on a curated Wikimedia-based Italian corpus \cite{wikimedia_it_dataset}.



Both English and Italian perplexity exhibited a consistent downward trend across training checkpoints, reflecting improved predictive accuracy of the underlying language model distribution. The concurrent reduction in perplexity and increase in MMLU accuracy provide complementary evidence that the model is simultaneously improving its token-level language modeling performance and its higher-level knowledge and reasoning capabilities.

%% file: Sections/05_long_context.tex
\clearpage
\section{Long-Context Adaptation}
As part of our ongoing model development efforts, we conducted a dedicated long‑context training phase aimed at significantly extending the effective context window of our language model. The primary objective was to enhance the model’s ability to capture long‑range dependencies, maintain coherence across multi‑page documents, and perform complex tasks such as document‑level summarization and multi‑step reasoning over extended inputs.

\begin{wrapfigure}{r}{0.5\textwidth}
    \centering
    \includegraphics[width=0.9\linewidth]{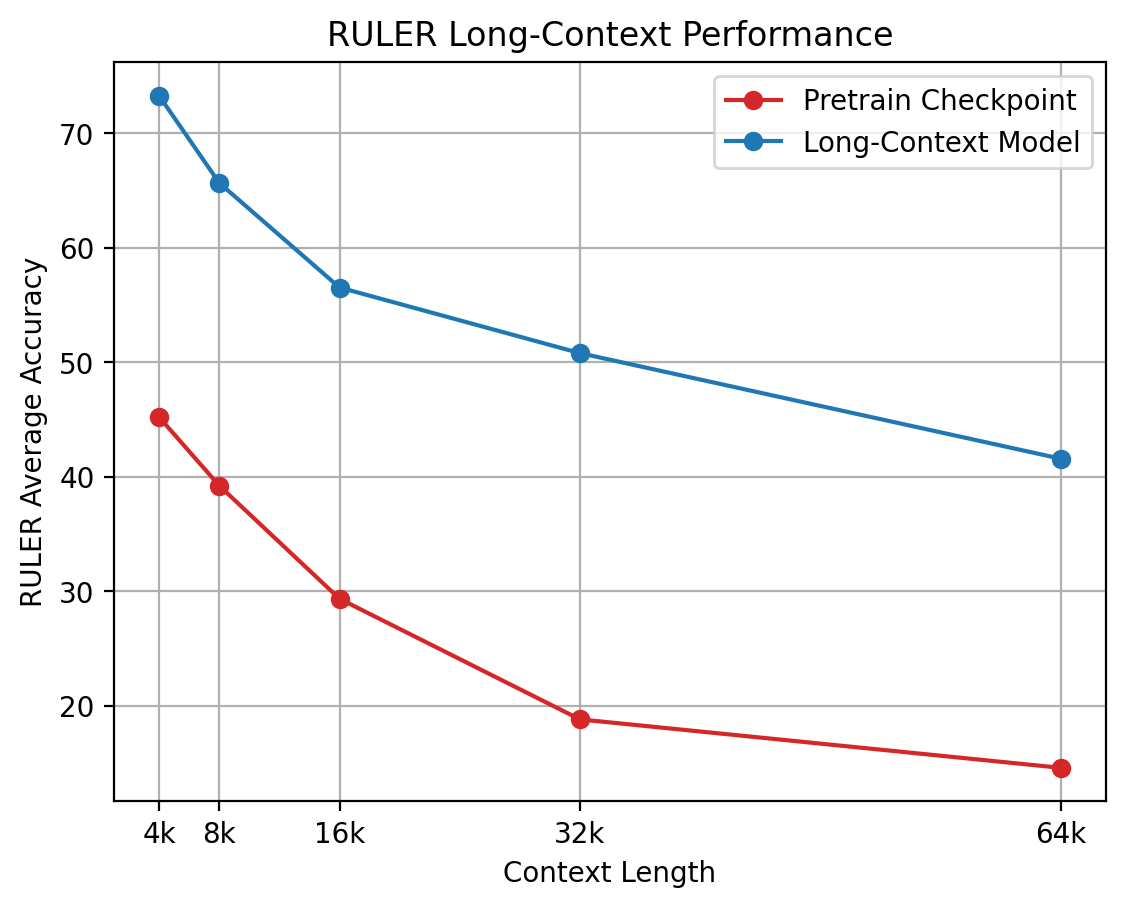}
    \caption{RULER long-context evaluation across increasing context lengths. Comparison between the final pre-training checkpoint (post stage 3) and the long-context-adapted model on the RULER benchmark. Both models are extended beyond 64k tokens using YARN to enable evaluation at long context lengths. Results are reported as average accuracy across the RULER task suite for context sizes ranging from 4k to 64k tokens.}
    \label{fig:long_context_eval}
\end{wrapfigure}

This long‑context adaptation initiative was designed to scale the model’s receptive field up to 65,536 tokens, enabling improved cross‑paragraph reasoning and more robust global information integration. To achieve this, we leveraged the Megatron‑LM framework, which provides optimized support for large‑scale transformer training through a combination of expert parallelism, pipeline parallelism, and context‑aware partitioning strategies.

A key focus of the adaptation was ensuring computational stability at very long sequence lengths. This required adjustments to positional encoding mechanisms and attention scaling strategies to preserve gradient quality and maintain throughput. Overall, the long‑context training stage strengthens the model’s performance in long‑form generation scenarios, increasing robustness and coherence across extended inputs.

\subsection{Long-Context Adaptation Data}
The allenai/dolmino-mix-1124 dataset is a large-scale English text corpus designed for advanced language model training \cite{dolmino_mix}. It contains a mixture of high-quality data sources—including web pages from DCLM, curated text like FLAN and Wikipedia dumps, STEM research content, StackExchange posts, and multiple synthetic mathematical subsets—totaling 50B tokens with diverse linguistic and factual content. The mixture strategy balances general web text with structured and domain-specific material, making it well suited for training long-context language models that benefit from both broad world knowledge and specialized reasoning samples. This dataset is released under an open data license (ODC-BY).

\subsection{Long-Context Adaptation Stage}
The long‑context adaptation stage introduced several architectural and training‑level configurations to support significantly extended sequence lengths while maintaining high training efficiency.

During this phase, the model was trained using a sequence length of 32,768 tokens, balancing memory feasibility with the progressive extension of long‑range attention capacity. Training employed a global batch size of 128, ensuring adequate gradient signal accumulation while preserving stability across distributed workers.

To enable efficient training at these sequence lengths, we introduced a context parallelism factor of 8, allowing the attention computation for long sequences to be partitioned across multiple devices without exceeding memory constraints. This context‑parallel setup was combined with the existing pipeline parallelism of 2 and expert parallelism of 4, achieving an effective distribution across compute nodes while maintaining high training throughput.

This adaptation phase was conducted on a subset of the full training corpus using 128 compute nodes (512 GPUs), covering roughly half of the total dataset due to time and budget constraints.

\subsection{Long-Context Adaptation Evaluation}
To assess the effectiveness of the long-context adaptation phase, we evaluate the model’s ability to process and reason over extended input sequences using the RULER benchmark \cite{hsieh2024ruler}. RULER provides a comprehensive framework for long-context evaluation, spanning 13 tasks grouped into four categories that probe retrieval, reasoning, aggregation, and robustness under increasing context lengths. 

We compare the final checkpoint obtained at the end of pre-training (post stage 3) with the model after the dedicated long-context adaptation phase. For both models, the context window is extended beyond 64k tokens using YARN \cite{peng2023yarn}, ensuring support for at least 64k-token sequences. The pre-training checkpoint is extended using a YARN scaling factor of 16, while the long-context-adapted model employs a YARN scaling factor of 2. 

Evaluation is conducted across multiple context lengths (4k, 8k, 16k, 32k and 64k tokens). Results, summarized in \autoref{fig:long_context_eval}, show that while the pre-trained model exhibits a sharp degradation in performance as context length increases, the long-context-adapted model maintains substantially higher accuracy across all evaluated settings.


These results indicate that positional extension alone, even when applied uniformly across models, is insufficient to preserve performance under long-context regimes. In contrast, targeted long-context adaptation leads to significantly improved robustness and scalability, enabling the model to effectively utilize extended context windows and mitigate the performance degradation observed in the pre-trained baseline. 

Despite the clear improvements over the pre‑training baseline, the overall performance remains relatively limited. It is likely that a more prolonged or extensive long‑context adaptation phase would have allowed the model to better consolidate the skills required for very long input sequences. In future work, we intend to extend this training phase to further enhance the model’s robustness and effectiveness in long‑context settings.

%% file: Sections/06_mid_training.tex
\section{Mid-Training}
The mid-training phase represents an intermediate consolidation step in the training pipeline, positioned between the initial large-scale training and the subsequent alignment and instruction-focused stages. Its goal is to refine the internal representations of the model under more controlled conditions, improving stability and preparing the model for later optimization stages, with a specific focus on reasoning structure.
Mid-training was conducted as a single large-scale run starting from a model checkpoint already supporting extended context lengths. Rather than introducing long-context capabilities, this phase aimed at consolidating and stabilizing the model behavior in this regime, preparing it for downstream alignment and instruction-tuning stages.
The training was executed using the Megatron-LM framework, selected for its native support for large-scale distributed training, expert parallelism, and efficient execution on HPC systems.

\subsection{Mid-Training Data}
The dataset used for mid-training consists of a compact mixture of publicly available post-training and structured reasoning corpora. The selection was designed to enable controlled model adaptation, with particular emphasis on conversational structure and explicitly represented reasoning traces. To preserve distributional consistency, only languages already present in the pre-training corpus were selected.
All samples were filtered to ensure correct language attribution, valid reasoning annotations, and consistent structural formatting. Special care was taken to enforce a clear separation between intermediate reasoning steps and final answers. Samples failing to meet the required structural or linguistic constraints were excluded.
Data preparation followed a unified internal pipeline that produced fully pre-templated and serialized training sequences using the project-specific chat template. The corpus is restricted to samples containing structured reasoning segments, in which assistant responses explicitly encode intermediate reasoning traces delimited by dedicated tags (e.g., \texttt{<think> ... </think>}).
This deterministic templating strategy directly materializes conversational context and reasoning structure within the training sequences. Such explicit formatting ensures stable separation between reasoning traces and final outputs, guarantees deterministic tokenization required by Megatron-LM, and maintains methodological consistency between mid-training data and subsequent reasoning-enabled evaluation.

Information about the datasets employed in the Mid-Training phase is summarized in \autoref{tab:mid_train_datasets}.

\begin{table}[h]
\centering
\setlength{\arrayrulewidth}{0.4pt}
\setlength{\tabcolsep}{15pt}
\renewcommand{\arraystretch}{2}

\begin{tabular}{c|c|c}
\textbf{Dataset} & \shortstack{\textbf{Licence}} & \shortstack{\textbf{Token Composition} \\ \textbf{(45B token)}} \\
\specialrule{1.5pt}{0pt}{0pt}

\shortstack{\textbf{Llama-Nemotron-Post-Training-Dataset\_reasoning\_r1}} 
& cc-by-4.0\tablefootnote{The dataset is largely open-licensed; samples with more restrictive terms were excluded from our training subset.} 
& 42\% \\

\textbf{OpenThoughts3-1.2M} 
& Apache-2.0 
& 12\% \\

\textbf{Dolci-ThinK-SFT-32B} 
& odc-by 
& 46\% \\

\end{tabular}

\vspace{5mm}
\caption{Mid-Train datasets}
\label{tab:mid_train_datasets}
\end{table}

\subsection{Mid-Training Stage}
Training was performed on 16 nodes (64 GPUs total), with a global batch size of 128 and a sequence length of 32,768 tokens. The training objective was defined in terms of total token budget, targeting approximately 100 billion tokens, corresponding to around 3 million training samples.
The run progressed for approximately 7 days, accumulating close to 25,000 training iterations, corresponding to an estimated compute usage of 12,000 GPU-hours. Loss values during this stage remained stable and well-behaved, converging from around 0.82 toward 0.74, while the auxiliary load-balancing loss remained close to 1.0, indicating stable expert utilization throughout the run. Memory usage and parameter norms remained constant, confirming numerical stability over extended training.
Multiple checkpoints were saved along the training trajectory. These checkpoints constitute the definitive mid-training outputs and are used both for subsequent evaluation and as initialization points for downstream alignment and instruction-tuning phases.

\subsection{Mid-Training Evaluation}
The evaluation of the mid-training phase is designed to assess the consolidation and stability of reasoning capabilities across checkpoints generated within the same optimization trajectory, rather than to measure absolute task performance or establish leaderboard-level results.
In contrast to pre-training evaluation, which tracks the emergence of general knowledge and foundational reasoning skills, mid-training evaluation focuses on verifying that continued optimization preserves previously acquired capabilities and improves reasoning consistency.

Evaluation was conducted on a selected set of intermediate checkpoints extracted from the mid-training run. These checkpoints represent different stages of the same optimization trajectory and enable controlled comparison of incremental capability consolidation. In line with the pre-training evaluation protocol, assessment continues to rely on MMLU accuracy as a representative benchmark for general knowledge and multi-domain reasoning. However, at this stage evaluation is performed in reasoning-enabled mode, activating structured reasoning traces through the mid training chat template.   This ensures continuity with earlier capability tracking while specifically monitoring the stabilization and consolidation of explicit reasoning behavior under extended-context conditions.

Results are reported in terms of relative deltas across checkpoints, emphasizing progressive consolidation rather than isolated peak performance.

\begin{table}[h]
\centering
\setlength{\arrayrulewidth}{0.4pt}
\setlength{\tabcolsep}{15pt}
\renewcommand{\arraystretch}{1.8}

\begin{tabular}{c|c|c}
\textbf{Step} & \textbf{MMLU} & \textbf{$\Delta$} \\
\specialrule{1.5pt}{0pt}{0pt}
12288 & 65.68 & +11.01 \\
16800 & 67.02 & +1.33 \\
20800 & 67.02 & +0.00 \\
23841 & 68.46 & +1.44 \\
\end{tabular}

\vspace{5mm}
\caption{MMLU accuracy across mid-training checkpoints.}
\label{tab:midtrain_mmlu}
\end{table}

The results reported in Table~\ref{tab:midtrain_mmlu} show a steady improvement in MMLU accuracy across checkpoints, indicating progressive consolidation of general reasoning capabilities. 
Performance increases consistently without late-stage degradation, suggesting stable optimization dynamics.

Overall, mid-training strengthens broad reasoning competence and provides stable checkpoints suitable for downstream alignment and instruction-tuning.

%% file: Sections/07_post_training.tex
\section{Post-Training}
The post‑training pipeline is organized as a coherent sequence of refinement steps that progressively consolidate the model’s instruction‑following capabilities, alignment quality, and functional performance.  Beginning from the selected mid‑training checkpoint, the pipeline combines supervised fine‑tuning, preference‑based optimization, and model souping. 

First, the model undergoes SFT to acquire strong instruction-following and conversational capabilities. Next, we apply Anchored Preference Optimization (APO) \cite{apo}, an off-policy variant of Direct Preference Optimization (DPO) \cite{rafailov2023direct}, following the approach adopted in SmolLM2 \cite{allal2025smollm2} and SmolLM3 \cite{huggingface2025smollm3}, to further align the model. Finally, we perform model merging through a model soup strategy, combining the SFT and aligned checkpoints to integrate their complementary strengths into a single, more robust and stable final model.

The core objective of post‑training was to produce a model with a well‑defined chat template and strong instruction‑following capabilities, while supporting both reasoning and non‑reasoning interaction modes. The model also provides native support for function calling compliant with the Model Context Protocol (MCP), ensuring structured, interoperable, and predictable tool‑usage behaviors.

For reasoning‑enabled interactions, the system supports both Italian and English. This multilingual reasoning capability is activated through dedicated control tokens embedded in the chat template, allowing explicit selection of the target reasoning language. In addition to standard chain‑of‑thought traces, the model introduces a synthetic, compressed reasoning modality (“turbo”), also controlled via dedicated tokens, designed to provide concise intermediate reasoning signals while reducing verbosity and computational overhead.

All post‑training stages were executed as full‑parameter optimization runs, utilizing between 4 and 32 compute nodes equipped with 4×A100 GPUs each. The training stack leveraged DeepSpeed ZeRO‑3 \cite{deepspeed1, deepspeed2} for efficient memory partitioning and large‑scale model parallelism, while the TRL library \cite{trl2020} was used to implement both supervised fine‑tuning and preference‑based optimization. Model souping was performed using mergekit \cite{mergekit}, integrated as the final consolidation step of the post‑training workflow.

\subsection{SFT Stage}
The first post-training stage consists of Supervised Fine-Tuning (SFT) for instruction alignment. In this phase, the model is trained on an instruction-following dataset, formatted using a custom conversational chat template. The template is inspired by the Qwen schema \cite{yang2025qwen3} and further adapted to match the target interaction style and system constraints.

This stage plays a foundational role in the overall pipeline, as it is where the final chat template is introduced and stabilized. Alongside standard instruction tuning, SFT is used to explicitly teach the model the structure and semantics of the conversational format, including system, user, and assistant roles, as well as the handling of control tokens.
Overall, the SFT stage focuses on grounding the model in high-quality instructional behavior, enforcing consistent response formatting, and ensuring robust adherence to user intent across both reasoning and non-reasoning interaction modes. The resulting SFT checkpoint serves as the initialization point for all subsequent post-training stages.

From an operational standpoint, the SFT phase was executed using a multi‑node distributed setup consisting of 4 compute nodes equipped with a total of 16 NVIDIA A100 GPUs. Training was performed over 5 full epochs with a gradient accumulation step (GA) set to 1, resulting in an effective global batch size (GBS) of 16. The optimization schedule followed a cosine‑decay learning‑rate policy, starting from an initial value of $2 \times 10^{-5}$  and progressively annealing to a minimum of $2 \times 10^{-6}$, which was reached toward the final portion of training.

\subsubsection{Reasoning Mode}
\label{subsec:reasoning_mode}
During the SFT stage, a set of dedicated control tokens is introduced to explicitly regulate reasoning behavior during training and inference. The model supports two inference regimes: non-reasoning mode and reasoning mode.

When reasoning is enabled (\texttt{enable\_thinking = true}), the reasoning configuration is controlled through the following tokens:
\begin{itemize}
    \item \textbf{Reasoning Structure}  
    
    Internal reasoning content is encapsulated between the tokens  \texttt{<think>} and \texttt{</think>}, which delimit intermediate reasoning from the final answer.
    
    \item \textbf{Language Control}  
    The opening \texttt{<think>} token is preceded by a reasoning language specifier:
    \begin{itemize}
        \item \texttt{/reasoning\_en}
        \item \texttt{/reasoning\_ita}
    \end{itemize}
    enabling reasoning traces in English or Italian.
    
    \item \textbf{Turbo Mode}
    
    When a synthetic and compressed reasoning mode is desired, an additional  \texttt{/turbo} token is inserted between the language token and the \texttt{<think>} tag. This activates a shorter and less verbose reasoning trace, reducing computational overhead while preserving the same structural format. 

\end{itemize}

When reasoning is disabled (\texttt{enable\_thinking = false}), the model preserves the structural tags \texttt{<think></think>} for formatting consistency, but does not generate any intermediate reasoning content.
Concrete template realizations for all supported reasoning configurations are provided in Appendix~\ref{app:reasoning_templates}.

By learning these tokens during supervised training, the model acquires explicit control over multiple reasoning configurations. This design enables the model to reliably distinguish between non-reasoning responses, full explicit reasoning traces in multiple languages, and concise intermediate reasoning outputs.

\subsubsection{SFT Training Data}
The Supervised Fine‑Tuning (SFT) dataset was constructed through a multi‑source, multi‑strategy pipeline aimed at achieving broad task coverage, diverse interaction patterns, and a balanced representation of reasoning behaviors, while preserving strong linguistic focus and sample heterogeneity. The final corpus contains approximately 2.5 billion tokens and consists of examples with variable lengths: most are below 4,096 tokens, whereas a smaller subset extends up to 32k tokens.

To prepare the data for training, all examples are organized using a 32k‑token packing scheme, bringing each training sequence close to the model’s maximum context length. When space permits, multiple independent examples are concatenated into a single packed sequence; individual examples are never split across different sequences. Attention masks ensure strict isolation between examples, preventing tokens from attending outside their own boundaries.
Within each packed sequence, the loss is computed only over regions containing valid training content. Tokens introduced by packing—such as padding, separators, and structural delimiters—are fully excluded from the loss through selective masking. This strategy improves token efficiency, minimizes padding overhead, and enables effective training on long‑context inputs while preserving the semantic and structural integrity of each example.



















\begin{table}[t]
\centering
\small
\setlength{\arrayrulewidth}{0.4pt}
\renewcommand{\arraystretch}{1.15}
\setlength{\tabcolsep}{4pt}

\begin{tabular}{p{3.2cm}|c|c|p{4.2cm}}

\textbf{Dataset} & \textbf{\#Tokens (M)} & \textbf{Mode} & \textbf{Task Type} \\
\noalign{\hrule height 0.8pt}

Policies Alignment & 5.3 & RI, RE, NR & Policies; Safety \\

Metrics Alignment & 31.4 & RE, NR & Metrics \\

General Purpose & 23.6 & RE, NR &
Creative Writing; Text Processing;Information Extraction; Structured Output; SQL-like Generation; Code Generation; Code Migration; Code Related Generation; ActLike; Data Mng; Agentic Orchestration; Misc \\

Mathematical Problems & 11.3 & RE &
Math \\

Reasoning Capabilities & 186.7 & RE &
Problem Solving; Code Generation; Code Related Generation \\

Italian Reasoning & 144.1 & RI &
Problem Solving; Code Generation; Code Related Generation \\

Turbo Reasoning & 8.3 & RTI, RTE &
Problem Solving; Code Generation; Code Related Generation \\

Tool Calling & 9.3 & RE, NR &
Tool Calling \\

Data Handling & 35.9 & NR, RE &
Data Management \\

Document Understanding & 0.2 & RE, NR &
Information Extraction \\

\end{tabular}

\vspace{5mm}
\caption{Training dataset mixture used for the SFT stage.}
\label{tab:sft_dataset}
\end{table}

The training datasets for all supported modalities were assembled using a hybrid data strategy combining heterogeneous instruction‑following sources, internally distilled datasets, and purpose‑built collections tailored to specific training objectives. To strengthen generalization and robustness, the mixture was designed to maximize diversity across data origins, task structures, and interaction patterns.

\autoref{tab:sft_dataset} summarizes the datasets included in the final Training Dataset mixture along with their token counts.A detailed taxonomy of the task categories included in the SFT dataset is provided in \autoref{app:sft_data_details}, which outlines the main task families and sub‑categories represented in the corpus.

A substantial fraction of data consists of instruction‑following datasets produced through distillation pipelines involving large external teacher models, notably Qwen3‑32B\cite{yang2025qwen3}, DeepSeek‑R1\cite{deepseekai2025} and GPT-OSS\cite{agarwal2025gpt}, whose permissive licenses explicitly allow model distillation and downstream reuse.
From each source, subsets of samples were selected using a diversity-oriented sampling strategy. Seed prompts were embedded and clustered, and samples were drawn across cluster centroids to favor semantic heterogeneity and reduce redundancy. 

From the same pool of reasoning datasets, and using an analogous clustering-based sampling approach, we constructed the \texttt{Italian Reasoning} dataset by selecting subsets of examples explicitly exhibiting active reasoning traces. For these samples, the original reasoning content was translated into Italian and then reinjected into the prompt to generate a new response conditioned on the translated reasoning content. The response generation and distillation step was performed using Qwen3-32B as the teacher model. This process increased the availability of high-quality Italian reasoning examples while preserving the logical structure of the original reasoning chains.

For a smaller but targeted subset of prompts, we constructed the \texttt{Turbo Reasoning} dataset by generating a synthetic compressed reasoning modality (``turbo''). Initially, only examples with reasoning in English were considered. In this setup, original prompts were used to distill concise reasoning traces using \texttt{gpt-oss} in low-reasoning mode. The resulting reasoning was intentionally structured in a bullet-point, highly compact format, enforced through prompt engineering constraints. Once the compressed reasoning and the corresponding conditioned responses were generated in English, the original prompts were revised by removing model-specific instructions. The English reasoning traces were then translated into Italian and reinjected into the prompts to generate new responses conditioned on the translated content, following the same approach used for \texttt{Italian Reasoning} dataset. The final \texttt{Turbo Reasoning} dataset therefore includes both the original English turbo reasoning traces and responses, as well as their Italian counterparts, directly injected into the standardized chat template. This dataset specifically targets efficient intermediate reasoning signals with reduced verbosity and computational overhead.

In addition to reasoning datasets, we constructed custom prompt datasets covering a broad range of instruction-following tasks, specifically \texttt{General Purpose} and \texttt{Policies Alignment}. These datasets were partly distilled from \texttt{gpt-oss-120B} as the teacher model and partly programmatically constructed to enforce specific alignment policies, ensuring that responses follow desired safety, fairness, and behavior guidelines. Prompts were manually designed and programmatically expanded to span multiple domains and interaction patterns, including the tasks listed in \autoref{tab:sft_dataset}. Prompt construction itself was supported by \texttt{gpt-oss-120B}, which was used to generate personas, domain-specific contexts, and relevant questions, ensuring variability in style, intent, and difficulty. The final datasets include both Italian and English samples, with a predominance of Italian and a minority of English examples, reflecting the target user distribution.

Finally, a set of datasets was derived from benchmark- and KPI-oriented tasks, forming the  \texttt{Metrics Alignment} dataset, and leveraging the prompting structures commonly used in evaluation harnesses. Whenever available, only training splits were used, with strict exclusion of any test data.

Specifically:
\begin{itemize}
    \item \textbf{AIME}: problems from editions spanning 1983 to 2023 were included.
    \item \textbf{MMLU}: questions from the training split were reformatted using the MMLU-Pro evaluation prompt structure.
    \item \textbf{GSM8K}: only training examples were used, following the standard evaluation-style prompting.
\end{itemize}

These datasets were explicitly designed to teach the model to produce structured, evaluation-compliant answers aligned with downstream benchmarking protocols, while carefully avoiding any contamination from held-out test sets.

\subsection{APO Stage}
Following the Supervised Fine-Tuning stage, the model was further refined through a preference-based optimization phase based on pairwise comparisons between alternative responses. This stage follows a Direct Preference Optimization (DPO) paradigm \cite{rafailov2023direct}, encouraging the model to assign higher probability to preferred (“chosen”) responses than to less desirable (“rejected”) ones, without introducing a separate reward model.

The optimization objective adopts an APO-based loss \cite{he2024accelerated}, which reformulates the preference objective in terms of relative advantages with respect to a reference model, rather than relying solely on raw likelihood differences between paired responses. 

This phase aims to improve response quality across multiple dimensions, including factual correctness, adherence to user intent, formatting consistency, and calibrated control over reasoning and non-reasoning modes.

Training is performed via full-parameter optimization initialized from the SFT checkpoint. While the Supervised Fine-Tuning phase was conducted with a maximum sequence length of 32k tokens, the subsequent preference optimization stage adopts a reduced context window of up to 16k tokens. This adjustment was introduced to mitigate memory constraints encountered during full-parameter training with paired preference samples, which effectively double the forward-pass memory footprint.

This stage was executed using a multi‑node distributed setup consisting of 32 compute nodes equipped with a total of 128 NVIDIA A100 GPUs. Training was performed over only one full epoch with a gradient accumulation step (GA) set to 1, resulting in an effective global batch size (GBS) of 128. The optimization schedule followed a cosine‑decay learning‑rate policy, starting from an initial value of  $5 \times 10^{-6}$ and progressively annealing to a minimum of  1\%.

A model souping strategy was employed to consolidate the benefits of multiple checkpoints along the training trajectory, following the general principle demonstrated in SmolLM3 \cite{huggingface2025smollm3}, where model merging is used to blend complementary behaviors emerging at different optimization stages without additional training cost. In our setup, we collected 16 checkpoints, saved at intervals of 100 optimization steps throughout the post training run, and used them to explore a wide range of merging configurations.

Starting from the initial SFT checkpoint, we constructed several candidate soups by combining it with subsets of 2 to 6 intermediate checkpoints. For each soup, we applied different weighting schemes to assess how the relative contribution of each checkpoint influenced the model’s performance. The tested configurations included:

\begin{itemize}
    \item Uniform low‑weight merges, where the SFT checkpoint dominated and intermediate checkpoints contributed with small, equal weights.
    \item Uniform high‑weight merges, giving more influence to the trajectory checkpoints while still preserving SFT as the reference anchors.
    \item Increasing‑weight merges, in which later checkpoints were assigned progressively larger weights to emphasize the most recent optimization signals.
    \item Decreasing‑weight merges, prioritizing earlier checkpoints that tended to preserve stability or earlier‑stage behaviors.
\end{itemize}

Each soup candidate was evaluated allowing us to empirically quantify how different merge recipes affected the overall alignment robustness. The model soups exhibited broadly similar evaluation results across the explored configuration space, indicating that the trajectory checkpoints contributed complementary but relatively stable improvements. However, a slight and consistent advantage was observed for one of the uniform low weight merge configurations, which was therefore selected as the final merged model.

\subsubsection{APO Training Data}
The APO training dataset is constructed by deriving preference pairs from a broad and heterogeneous collection of tasks, domains, and response styles. To ensure continuity and avoid distributional drift across post-training stages, all APO inputs are sourced directly from the original SFT corpus. The same preprocessing, formatting, and filtering pipeline is applied, guaranteeing that prompt structures, instruction styles, and conversational patterns remain aligned between SFT and preference optimization.

For each SFT sample, dedicated input prompts are extracted to serve as seeds for the subsequent generation of candidate responses used in APO.

\begin{itemize}
    \item \textbf{Single-turn samples:} the APO input consists of the original system message and the user message exactly as present in the SFT corpus.
    
    \item \textbf{Multi-turn samples:} a separate APO input is created for each assistant turn. For a conversation containing $n$ assistant messages, $n$ APO samples are produced, each containing the full conversation history up to the assistant turn for which a response must be regenerated. This ensures that the APO model learns to optimize preferences across realistic multi-turn interaction contexts while preserving the full conversational dependency structure.
\end{itemize}

Preference pairs are generated through a controlled distillation setup inspired by the approach adopted in SmolLM3 \cite{huggingface2025smollm3}: candidate responses are produced using a dual-model scheme involving Qwen3-32B and Qwen3-0.6B, yielding paired outputs spanning both reasoning-enabled and concise answer formats. This setup provides a consistent, well-structured supervision signal across reasoning and non-reasoning modes, while maintaining strict control over the qualitative properties of the preference direction used during optimization.

Samples span a broad spectrum of context lengths: the majority remain under 4k tokens, while progressively smaller fractions extend to 8k and up to 16k tokens. Overall, the constructed preference dataset comprises roughly one billion tokens.

\subsection{Post-Training Evaluation}
A detailed evaluation of the final model is provided in the \autoref{sec:benchmarking}, which includes the complete benchmark results and comparative analyses.

Here, we report only a concise snapshot of the model's progression across the post‑training pipeline, using a small set of representative KPIs to illustrate the evolution from mid‑training to the SFT and APO stages.
As shown in \autoref{tab:post_train_eval}, the SFT stage yields only marginal improvements on broad knowledge benchmarks such as MMLU-Pro and MMLU‑Redux, while producing a substantial gain on AIME, suggesting that supervised instruction tuning reinforces structured reasoning traces more directly. The most pronounced improvement occurs during the APO stage, which delivers meaningful gains across all reported metrics, particularly on reasoning‑heavy tasks.

\begin{table}[h]
\centering
\setlength{\arrayrulewidth}{0.4pt}
\setlength{\tabcolsep}{15pt}
\renewcommand{\arraystretch}{1.2}

\begin{tabular}{c|c|c|c}
\textbf{Stage} 
& \textbf{MMLU-Pro}  
& \textbf{MMLU-Redux}  
& \textbf{AIME26}  \\
\specialrule{1.5pt}{0pt}{0pt}

Mid-Training & 50.4 & 66.8 & 47 \\
SFT Stage & 50.7 & 73.2 & 53 \\
APO Stage & 57.3 & 75.5 & 70 \\
\end{tabular}

\vspace{5mm}
\caption{KPIs across post-training stages.}
\label{tab:post_train_eval}
\end{table}

%% file: Sections/08_evaluation.tex
\clearpage
\section{Final Benchmarking and Comparative Analysis}
\label{sec:benchmarking}
This chapter presents the final benchmarking results for the fully post‑trained model and compares its performance against a set of reference models of comparable and larger size. The evaluation focuses on core capability areas—including reasoning, instruction following, structured output generation, multilingual competence, and function calling—to provide a comprehensive view of the model’s strengths and limitations.
The methodology and KPI selection follow the approach detailed in \autoref{sec:eval_framework}, which describes the evaluation framework, the metrics adopted, and the criteria used to ensure consistency across tasks and benchmarks.

\subsection{Model Comparisons}
The evaluation scores obtained under the unified generation setup are reported in \autoref{tab:eval_results_standard_parameters}, while results derived from each model’s optimal configuration are presented in \autoref{tab:eval_results_opt_parameters}. In addition, for a focused view on Italian‑specific performance, \autoref{tab:ita_results} provides a consolidated comparison across both standardized and optimal evaluation setups. Together, these two perspectives provide complementary insights: the standardized setting enforces strict cross‑model comparability, whereas the optimal‑configuration results capture the maximum practical performance achievable by each system under realistic serving constraints. Across both evaluation regimes, EngGPT2‑16B‑A3B consistently ranks among the strongest models within the comparable‑compute group, showing particularly strong gains on reasoning‑oriented benchmarks such as MMLU‑Pro and AIME. Moreover, despite its lower effective compute cost, EngGPT2‑16B‑A3B remains competitively positioned relative to substantially larger dense models, exhibiting only a modest performance gap overall.


At the same time, the model shows below‑expected performance on coding‑focused tasks (HumanEval) and tool‑use benchmarks (BFCLv3), which marginally lowers its aggregate score. We attribute this behavior to a relative underrepresentation of coding and tool‑interaction data during the SFT phase, an aspect we plan to reinforce in upcoming releases. Addressing this imbalance is expected to further improve the model’s robustness and close the remaining gap in these capabilities.

\begin{table}[h]
\centering
\setlength{\arrayrulewidth}{0.4pt}
\setlength{\tabcolsep}{4pt}
\renewcommand{\arraystretch}{1.6}
\small

\resizebox{\textwidth}{!}{%
\begin{tabular}{c|c|c|c|c|c|c|c|c|c}

\textbf{Group} &
\textbf{Model} &
\textbf{MMLU-Pro} &
\textbf{MMLU-Redux} &
\textbf{IFEval} &
\textbf{HumanEval} &
\textbf{AIME25} &
\textbf{AIME26} &
\textbf{GSM8K} &
\textbf{BFCL} \\
\cline{1-10}

\multirow{4}{*}{\textbf{Comparable}} &
\textbf{EngGPT2-16B-A3B} & \textbf{57.3} & \textbf{75.5} & \textbf{72} & \textbf{64} & \textbf{60} & \textbf{70} & \textbf{88} & \textbf{48.5}  \\
& Moonlight-16B-A3B-Instruct & 47.1 & 70.1 & 51.4 & 77.4 & 20.0 & 10.0   & 83.8 & 42.2 \\
& Llama-3.1-8B-Instruct      & 49.3 & 68.8 & 79.5 & 73.8 & 10.0  & 0.1   & 82.7 & 28.9 \\
& gemma-2-9b-it              & 49.2 & 74.7 & 75.2 & --   & 0.0    & 0.0    & 82.3 & 46.3 \\
\cline{1-10}

\multirow{5}{*}{\textbf{Larger}} &
Gpt-oss-20b-high           & 75.2 & 89.6 & 86.8 & 94.5 & 80   & 86.7   & 93.3 & 60.9\\
& Qwen3-14B                  & 77.3 & 89.9 & 90.4 & 97   & 83.3 & 86.7 & 93 & 74.3 \\
& Qwen3-30B-A3B              & 78.6 & 90.0   & 91.2 & 96.3 & 86.7 & 86.7 & 92.2 & 74.1 \\
& gemma-3-12b-it             & 57.6 & 75.3 & 82.4 & --   & 33.3 & 36.7 & 88.6 & 52.2 \\
\end{tabular}}
\vspace{5mm}
\caption{Benchmark results obtained using a standardized generation configuration across all evaluated models to ensure strict cross-model comparability.}
\label{tab:eval_results_standard_parameters}
\end{table}

\begin{table}[h]
\centering
\setlength{\arrayrulewidth}{0.4pt}
\setlength{\tabcolsep}{4pt}
\renewcommand{\arraystretch}{1.6}
\small

\resizebox{\textwidth}{!}{%
\begin{tabular}{c|c|c|c|c|c|c|c|c|c}
\textbf{Group} &
\textbf{Model} &
\textbf{MMLU-Pro} &
\textbf{MMLU-Redux} &
\textbf{IFEval} &
\textbf{HumanEval} &
\textbf{AIME25} &
\textbf{AIME26} &
\textbf{GSM8K} &
\textbf{BFCL} \\
\hline

\multirow{4}{*}{\textbf{Comparable}} &
\textbf{EngGPT2-16B-A3B} & \textbf{57.3} & \textbf{75.5} & \textbf{72.0} & \textbf{64.0} &  \textbf{60.0} & \textbf{70.0} & \textbf{88.0} & \textbf{48.5} \\
& Moonlight-16B-A3B-Instruct & 47.1 & 70.1 & 51.4 & 77.4 & 20.0 & 10.0 & 83.8 & 42.2 \\
& Llama-3.1-8B-Instruct      & 49.2 & 68.4 & 81.9 & 75.0  & 3.3 & 0.0 & 82.7 & 37.4 \\
& gemma-2-9b-it              & 48.2 & 74.8 & 72.0 & --   & 0.0    & 0.0    & 82.3 & 44.4 \\
\hline

\multirow{5}{*}{\textbf{Larger}} &
Gpt-oss-20b-high            & 63.1 & 89.8 & 91.1 & 98.2 & 73.3 &  76.7 & 93.9 & 58.9 \\
& Qwen3-14B                  & 77.3 & 89.9 & 90.4 & 97   & 83.3 & 86.7 &93.0 & 74.3 \\
& Qwen3-30B-A3B              & 78.6 & 90.0 & 91.2 & 96.3 & 86.7 & 86.7 & 92.2 & 74.1 \\
& gemma-3-12b-it             & 56.0 & 75.2 & 82.8 & --   & 30.0 & 33.3 & 88.6 & 52.2 \\

\end{tabular}}
\vspace{5mm}
\caption{Comparison among EngGPT2-16B-A3B and other comparable models, each evaluated under its optimal serving and decoding configuration (see \autoref{app:eval_config}), reflecting maximum achievable performance.}
\label{tab:eval_results_opt_parameters}
\end{table}

\begin{table}[h]
\centering
\setlength{\arrayrulewidth}{0.4pt}
\setlength{\tabcolsep}{6pt}
\renewcommand{\arraystretch}{1.25}
\small

\begin{tabular}{c|c|c|c|c|c}
\textbf{Group} & \textbf{Model} &
\shortstack{\textbf{ARC-IT} \\ \textit{Standard-config}} &
\shortstack{\textbf{MMLU-IT} \\ \textit{Standard-config}} &
\shortstack{\textbf{ARC-IT} \\ \textit{Optimal-config}} &
\shortstack{\textbf{MMLU-IT} \\ \textit{Optimal-config}} \\
\hline

\multirow{4}{*}{\textbf{Comparable}}
& \textbf{EngGPT2-16B-A3B} & \textbf{85.6} & \textbf{65.5} & \textbf{85.6} & \textbf{65.5} \\
& Moonlight-16B-A3B-Instruct & 63.1 & 49.2 & 63.1 & 49.2 \\
& Llama-3.1-8B-Instruct      & 81.4 & 60.6 & 80.0 & 60.6 \\
& gemma-2-9b-it              & 88.4 & 67.2 & 87.6 & 66.5 \\
\hline

\multirow{5}{*}{\textbf{Larger}}
& Gpt-oss-20b-high     & 92.6 & 80.2 & 93.2 & 80.4 \\
& Qwen3-14B            & 89.2 & 77.6 & 89.2 & 77.6 \\
& Qwen3-30B-A3B        & 95.0 & 81.5 & 95.0 & 81.5 \\
& gemma-3-12b-it       & 88.5 & 68.5 & 88.1 & 69.0 \\
\end{tabular}

\vspace{5mm}
\caption{Comparison between EngGPT2‑16B‑A3B and other comparable models on Italian benchmarks, evaluated under both standardized and optimal configurations (see \autoref{app:eval_config}).}
\label{tab:ita_results}
\end{table}

\begin{figure}[h]
    \centering
    \includegraphics[width=0.95\textwidth]{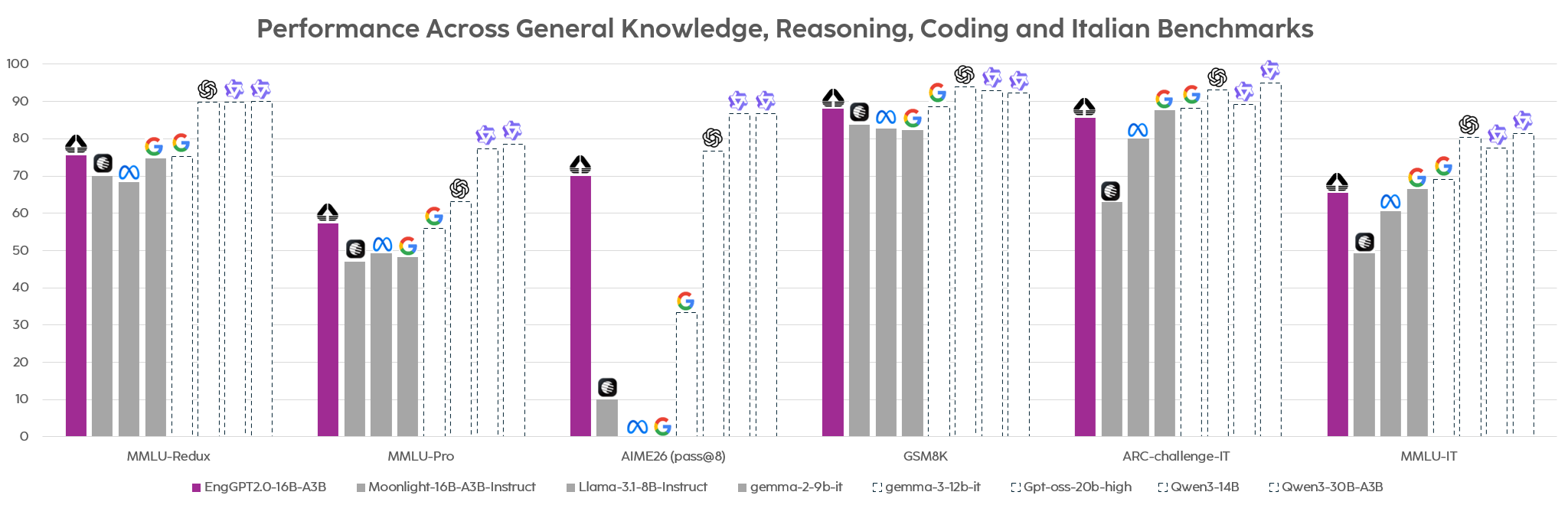}
    \caption{Visual comparison of general knowledge, reasoning and Italian benchmarks across EngGPT2-16B-A3B and representative open-weight models.}
    \label{fig:benchmark_visual_comparison}
\end{figure}

In addition to the comprehensive numerical comparison reported in \autoref{tab:eval_results_opt_parameters}, we provide a visual summary over a selected subset of benchmarks (see \autoref{fig:benchmark_visual_comparison}) to facilitate qualitative comparison across models. The selected benchmarks cover general knowledge, reasoning, and Italian-language evaluation tasks, offering a balanced representation of the main capability dimensions discussed in this section.

In addition to the numerical comparison in  \autoref{tab:eval_results_opt_parameters}, we also provide a visual overview on a selected subset of models and benchmarks (see \autoref{fig:benchmark_visual_comparison}), spanning general‑knowledge, reasoning, and Italian‑language tasks to offer a concise qualitative perspective across systems.

Overall, this combined tabular and visual analysis confirms that EngGPT2-16B-A3B outperforms several dense models in the 8B–12B range in reasoning capabilities, remains competitive with state-of-the-art open-weight models of comparable effective scale, and significantly narrows the gap with the strongest models in the 14B–30B class.

The results obtained with the standard configuration and the optimal configuration are closely aligned, depicting a consistent scenario with only minor variations. Therefore, all subsequent analyses will be conducted based on the optimal configuration.

\subsection{Normalization and Results}

While \autoref{tab:eval_results_opt_parameters} reports absolute benchmark scores under each model’s optimal serving configuration, direct comparison across heterogeneous benchmarks and models can be misleading due to differences in scale, task difficulty, and metric interpretation. To obtain a scale-independent aggregate indicator under strictly controlled evaluation conditions, we derive a \textit{Normalized Mean KPI} starting from the benchmark results reported in \autoref{tab:eval_results_opt_parameters}.

First, we perform a row-wise normalization. For each benchmark $b$, the highest-performing model is assigned a value of 100, and all other models are linearly rescaled relative to that maximum:

\[
\text{KPI}_{\text{norm}}(m,b) =
\frac{\text{Score}(m,b)}
{\max_{m'} \text{Score}(m',b)} \times 100
\]

where $\text{Score}(m,b)$ denotes the raw benchmark score of model $m$ on benchmark $b$. This procedure ensures that each benchmark contributes equally to the aggregate indicator, independently of its original numerical range.

We then compute, for each model $m$, the arithmetic mean of its normalized benchmark scores:

\[
\text{MeanKPI}_{\text{norm}}(m) =
\frac{1}{|B_m|}
\sum_{b \in B_m}
\text{KPI}_{\text{norm}}(m,b)
\]

where $B_m$ denotes the set of benchmarks with valid entries for model $m$. Missing values are excluded from the computation.

This two-step procedure yields a single composite indicator per model, reflecting its overall relative performance across the benchmark suite while avoiding distortions introduced by heterogeneous metric scales.

\begin{figure}[h]
\centering
\includegraphics[width=\linewidth]{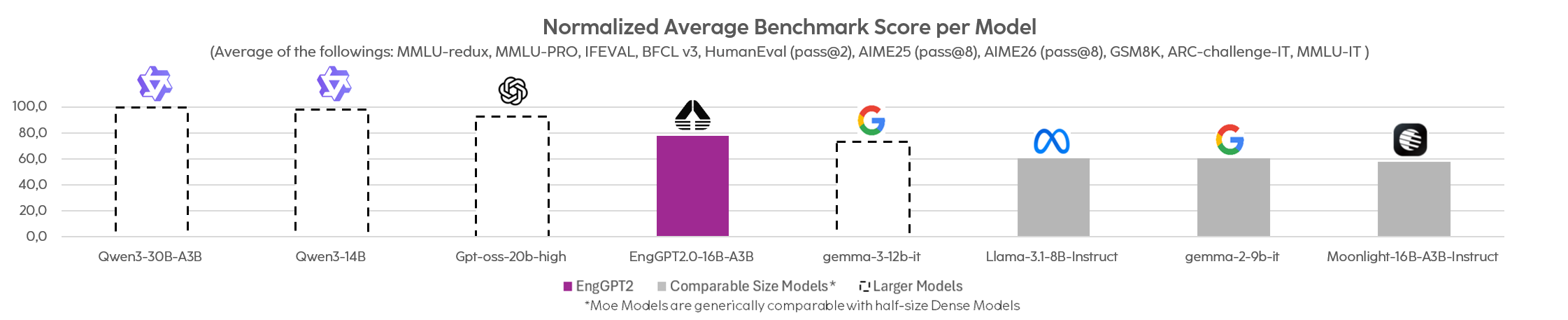}
\caption{Normalized average benchmark score per model, derived from row-wise normalized benchmark results.}
\label{fig:normalized_kpi}
\end{figure}

In normalized terms, the results reported in \autoref{fig:normalized_kpi} clearly reinforce the observations discussed in the preceding sections. Once benchmark scales are equalized, the model continues to emerge as the strongest performer among comparable architectures, consistently achieving the highest aggregated KPI across the evaluation suite. Moreover, the performance gap relative to Gemma‑3‑12‑it narrows substantially under this normalization procedure, indicating that their capabilities are effectively on par. Even when compared to significantly larger models, the relative shortfall remains limited, highlighting the model’s strong efficiency–performance trade‑off despite its smaller effective compute footprint.

To contextualize performance with respect to architectural scale and training investment, we additionally relate the normalized performance indicator to total training tokens and active parameters at inference time. Active parameters correspond to the effective compute footprint per forward pass and are particularly relevant for Mixture-of-Experts architectures, where only a subset of parameters is activated per token.

We define training and inference efficiency as:

\[
\text{TrainingEfficiency}(m) =
\frac{\text{MeanKPI}_{\text{norm}}(m)}{\text{TrainingTokens}(m)}
\]

\[
\text{InferenceEfficiency}(m) =
\frac{\text{MeanKPI}_{\text{norm}}(m)}{\text{ActiveParams}(m)}
\]

\begin{figure}[h]
\centering
\includegraphics[width=\linewidth]{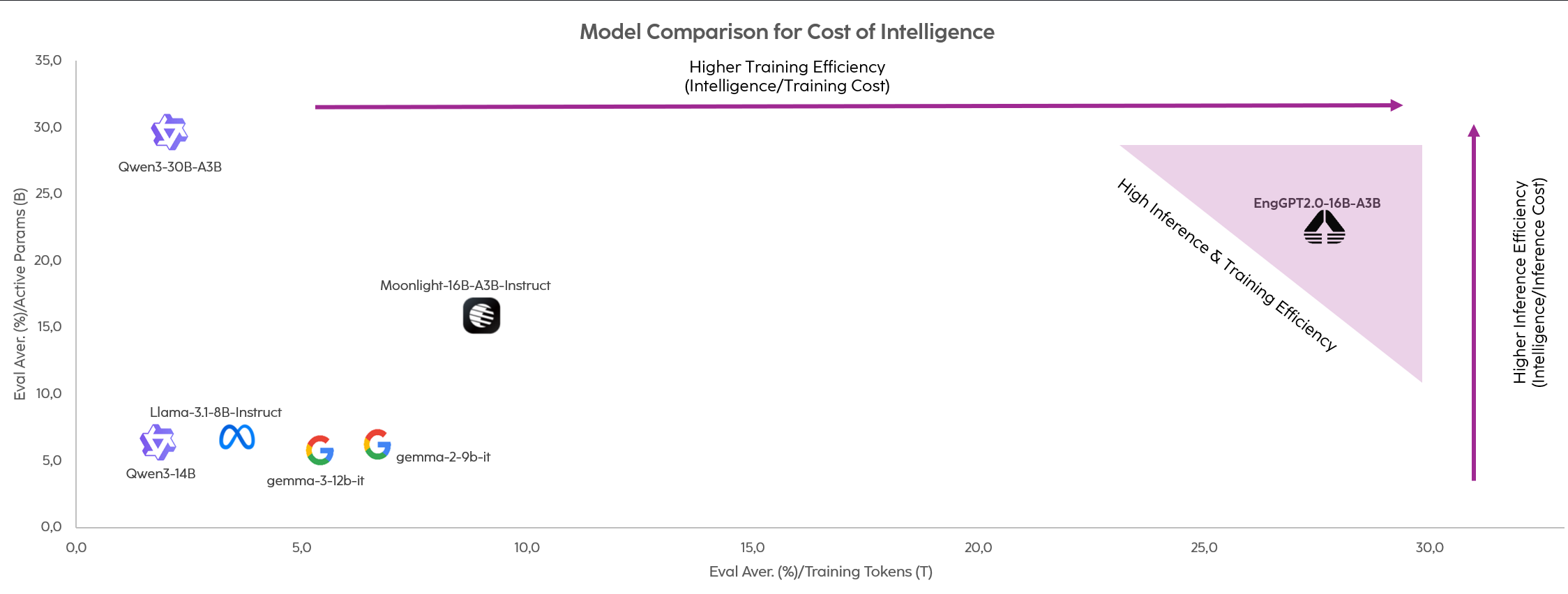}
\caption{Model comparison under the Cost of Intelligence framework. The horizontal axis represents training efficiency (normalized performance per training token), while the vertical axis represents inference efficiency (normalized performance per active parameter). Models positioned toward the upper-right region exhibit higher overall efficiency.}
\label{fig:cost_of_intelligence}
\end{figure}

When examining the normalized KPIs in relation to both training cost and inference‑time compute, the patterns shown in \autoref{fig:cost_of_intelligence} become even more striking. The lower‑left quadrant is dominated by dense models trained on very large token budgets, which naturally inflate overall training cost while offering limited efficiency in terms of performance per unit of compute.
Against this backdrop, Qwen3‑30B‑A3B stands out in the upper‑left quadrant: despite activating only 3B parameters per token at inference time—yielding top‑tier KPI performance at an exceptionally low per‑step compute footprint—its position also reflects a very high training cost, driven by its 36T training tokens. 
In contrast, EngGPT 2 is the only model positioned firmly in the upper‑right region, indicating a uniquely advantageous efficiency profile. It delivers strong normalized performance per training token and per active parameter, achieving a balanced and favorable trade‑off across both axes. This makes EngGPT 2 the model that best combines competitive capability with a substantially more economical training procedure and a cost‑effective serving footprint.

\subsection{Performance Comparison across Reasoning Modes}
The final model exposes multiple inference modes designed to support different interaction and deployment requirements. In addition to direct-answer generation (without explicit reasoning), the system supports full structured reasoning through dedicated control tokens, as well as a compressed reasoning modality (\textit{turbo}) that produces significantly shorter intermediate reasoning traces. Both reasoning modalities are available in English and Italian through the control-token mechanism introduced in Section~\ref{subsec:reasoning_mode}. These configurations do not alter the underlying model parameters but affect how the model externalizes its reasoning process at inference time. 

From a deployment perspective, these modes are relevant because generation length directly influences latency and token usage, which in turn impact inference cost in real-world serving environments. For this reason, we evaluated the model across the different inference configurations to assess how reasoning format influences benchmark performance across knowledge, reasoning, and multilingual tasks

\begin{table}[h]
\centering
\caption{Reasoning Modalities Evaluation}
\label{tab:reasoning_modes}
\begin{tabular}{lcccc}
\toprule
\textbf{Benchmark} & \textbf{Reasoning (EN)} & \textbf{Reasoning (ITA)} & \textbf{Turbo (EN)} & \textbf{Turbo (ITA)} \\
\midrule
MMLU-Pro          & 57.3 & 57.2 & 41.3 & 40.9 \\
MMLU-Redux        & 75.5 & 68.0 & 62.6 & 62.1 \\
AIME25            & 60.0 & 60.0 & 16.7 & 30.0 \\
AIME26            & 70.0 & 63.2 & 10.0 & 23.0 \\
GSM8K             & 88.0 & 88.0 & 74.4 & 75.8 \\
\bottomrule
\end{tabular}
\end{table}

\begin{figure}[h]
\centering
\includegraphics[width=0.75\linewidth]{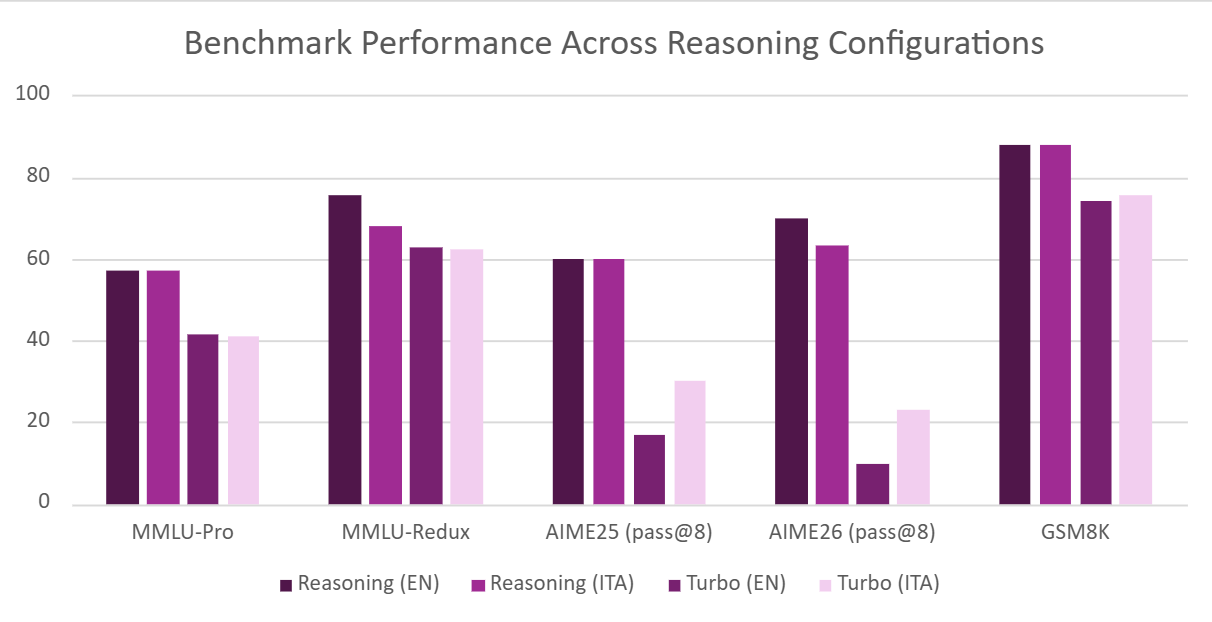}
\caption{Benchmark performance across full and compressed reasoning configurations in English and Italian.}
\label{fig:reasoning_modes}
\end{figure}

The performance patterns reported in Table~\ref{tab:reasoning_modes} are visually summarized in Figure~\ref{fig:reasoning_modes}.
Across benchmarks, performance trends are consistent with the structure of the training pipeline. During mid-training, structured reasoning traces were introduced primarily in English. In the subsequent post-training phase, English reasoning was further consolidated, and additional reasoning variants were introduced, including Italian reasoning and compressed (turbo) configurations.

Full English reasoning represents the most stable configuration overall, indicating that structured reasoning capabilities were most extensively reinforced in that setting. Italian reasoning closely follows, with limited task-dependent variation, suggesting that post-training successfully aligned reasoning behavior across languages under the standard (non-compressed) regime.

Turbo reasoning, introduced as a compressed variant during post-training, exhibits a systematic reduction in performance relative to full reasoning. This behavior is expected, as compression constrains the depth and verbosity of intermediate reasoning traces. The impact of compression depends on task characteristics: tasks requiring deep multi-step symbolic consistency are more sensitive to reduced reasoning depth, whereas knowledge-oriented or shorter-chain reasoning workloads exhibit more contained degradation.

Observed variations between English and Italian turbo configurations should be interpreted cautiously. Differences are more plausibly driven by task complexity and generation stochasticity under compressed reasoning than by systematic cross-lingual instability.

Language conditioning behaves reliably in full reasoning mode: when a reasoning language is explicitly selected, intermediate reasoning traces are consistently generated in the requested language. However, empirical inspection reveals a configuration-specific limitation in turbo mode. Although structural tokens and formatting are correctly enforced (e.g., \texttt{/reasoning\_ita /turbo <think>}), the internal reasoning language may follow the language of the user prompt rather than the explicitly selected \texttt{reasoning\_lang} parameter. In particular, under compressed reasoning, the model tends to align the reasoning language with the prompt language, even when a different reasoning language is specified at inference time.

This indicates that, in turbo mode, prompt-level language signals can dominate over auxiliary control tokens. As a result, structural compliance is preserved, but semantic language control behaves as a soft conditioning mechanism rather than a strict constraint. Importantly, this phenomenon appears to be specific to turbo reasoning and is not observed in full‑reasoning mode. A plausible explanation is that the turbo‑reasoning component was fine‑tuned on a comparatively smaller SFT dataset than the one used for full reasoning, which may limit its robustness and lead to the observed behavior. When the prompt language and the requested reasoning language are aligned, behavior remains stable and predictable. 


\begin{table}[h]
\centering
\caption{Reasoning Efficiency Trade-off}
\label{tab:reasoning_efficiency}
\begin{tabular}{lcccc}
\toprule
\textbf{Benchmark} & \textbf{Avg Tokens (Reasoning)} & \textbf{Avg Tokens (Turbo)} & \textbf{Perf. Drop (\%)} & \textbf{Tokens Drop (\%)} \\
\midrule
MMLU-Pro   & 2990  & 249  & 16.0 & 91.67 \\
MMLU-Redux & 1245  & 132  & 12.9 & 89.40 \\
AIME25     & 20061 & 746  & 43.3 & 96.28 \\
AIME26     & 19590 & 1753 & 60.0 & 91.05 \\
GSM8K      & 731   & 122  & 13.6 & 83.31 \\
\bottomrule
\end{tabular}
\end{table}

\autoref{tab:reasoning_efficiency} quantifies the efficiency trade-off introduced by compressed reasoning. Across all benchmarks, turbo mode substantially reduces the number of generated reasoning tokens, with average reductions ranging from approximately 83\% to over 96\% relative to full reasoning traces. This compression translates into significantly lower inference cost and latency, particularly on tasks where full reasoning produces long intermediate chains.

The performance impact of this reduction varies depending on the complexity of the benchmark. Knowledge-oriented and shorter reasoning tasks such as GSM8K and MMLU-Redux exhibit relatively moderate performance degradation (13--16\%), indicating that many reasoning steps can be compressed without fully compromising answer accuracy. In contrast, tasks requiring deeper symbolic reasoning, such as the AIME benchmarks, experience a larger performance drop under turbo mode. In these settings, full reasoning traces can exceed 20k tokens on average, and aggressive compression limits the model's ability to externalize long multi-step reasoning chains.

Overall, the results indicate that turbo reasoning substantially reduces the number of generated reasoning tokens across benchmarks, typically by close to an order of magnitude, while retaining reasonable performance on several tasks. These observations suggest that compressed reasoning can represent a practical deployment option when inference cost or latency are primary considerations, whereas full reasoning remains preferable for workloads that require deeper and more reliable multi-step reasoning.

%% file: Sections/09_conclusion.tex
\section{Conclusion}
In this technical report, we have presented EngGPT 2, a Mixture‑of‑Experts language model featuring 16B total parameters and 3B active per token. The architecture is designed to maximize compute, training, and inference efficiency, delivering competitive performance while operating with a fraction of the data and active parameters used by larger international models. Despite being trained on a comparatively modest corpus of 2.5T tokens, EngGPT 2 consistently matches or surpasses dense models in the 8B–16B parameter class across a broad range of benchmarks, with even stronger relative performance when normalized by training tokens or active inference compute.

The model supports multiple inference modes, including standard non‑reasoning responses, full structured reasoning in Italian and English, and a compressed “turbo” reasoning modality optimized for low‑latency, low‑verbosity applications. In addition, EngGPT 2 incorporates native tool‑calling aligned with modern interoperability standards and uses a carefully engineered chat template that ensures deterministic formatting, explicit control signals, and reliable behavior across interaction modes.

Evaluation results confirm that EngGPT 2 achieves solid performance in general knowledge (MMLU‑Pro, MMLU‑Redux), mathematical reasoning (AIME, GSM8K), code generation (HumanEval), and multilingual tasks (MMLU‑IT, ARC‑Challenge‑IT). In Italian‑centric evaluation in particular, EngGPT 2 establishes a new competitive baseline among open‑weight European models of comparable effective scale. Normalized analyses highlight what we consider the principal strength of this model family: high capability per unit of compute, demonstrating that sparse architectures and efficient training pipelines can serve as robust foundations for scalable and sustainable European LLM development.

EngGPT 2 represents a significant step toward a family of open, efficient, and regulation‑aligned European large language models. With this work, we aim to contribute to a broader ecosystem of trustworthy AI systems developed according to European values and operational requirements, and to establish a strong, sustainable foundation for future advancements.

Looking ahead, we acknowledge that the model’s performance on long‑context tasks remains limited and would have benefited from a more extended training schedule. Similarly, the Supervised Fine‑Tuning phase was likely too short and disproportionately focused on reasoning capabilities, resulting in comparatively weaker performance in tool calling and coding. Strengthening these two areas will therefore be a primary objective for upcoming releases. In parallel, we intend to scale both the size of the model and the volume of training tokens to further enhance overall capacity and robustness. We are also actively experimenting with RLVR, with the expectation that an additional reinforcement‑learning stage may yield further gains in reasoning stability, controllability, and downstream task performance.

%% file: Sections/10_Legal_notice.tex
\clearpage
\section{Legal Notice}
© Engineering Ingegneria Informatica S.p.A. All rights reserved.

This document is made publicly available for informational and research purposes. The content may be cited, provided that proper attribution is given to Engineering Ingegneria Informatica S.p.A.

No part of this document may be reproduced, modified, or distributed for commercial purposes without prior written authorization from Engineering Ingegneria Informatica S.p.A., unless otherwise explicitly permitted.

This document is provided “as is” and for informational purposes only. Engineering Ingegneria Informatica S.p.A. makes no representations or warranties, express or implied, regarding the accuracy, completeness, or fitness for a particular purpose of the information contained herein.

The content of this document does not constitute any contractual obligation or commitment.

%% file: Sections/11_appendixA.tex
\section{Details on Pretraining Datasets}
\label{appendix:datasets}

This appendix summarizes the datasets used for pre-training, describing their sources, scale, licensing conditions, and the data curation pipelines adopted by the original dataset authors.

\subsection{FineWeb-2}
FineWeb-2 \cite{huggingfacefw_2024} is sourced from 96 snapshots of Common Crawl collected between summer 2013 and April 2024, comprising approximately 8TB of compressed data, corresponding to roughly 3 trillion tokens across 5 billion documents in over 1,000 languages. For our experiments, we selected five Romance and Germanic languages: German, French, Italian, Portuguese, and Spanish. The dataset is released under the Open Data Commons Attribution (ODC-By) license, which permits both research and commercial use subject to the Terms of Use of Common Crawl. The raw web data was processed to extract and filter text from Common Crawl WARC files, with language-specific deduplication and quality filters applied to ensure high-quality training data. For all the FineWeb-related data, the processing pipeline includes anonymization of e-mail addresses and public IP addresses. However, given the web-sourced nature of the dataset, personally identifiable information such as names and phone numbers may still be present, and the authors provide an opt-out form for PII removal requests.

\subsection{FineWeb-Edu}

We incorporated educational content from FineWeb-Edu \cite{huggingfacefwedu_2024}, a curated subset of high-quality educational webpages. FineWeb-Edu is derived from FineWeb through an educational quality classification pipeline: a regression model was trained on 450,000 annotations from Llama3-70B-Instruct assigning educational scores from 0 to 5 to web documents. Applying a threshold of 3, this filtering procedure retained 1.3 trillion tokens while removing approximately 92\% of the original dataset. We selected data from Common Crawl snapshots CC-MAIN-2021 through CC-MAIN-2025, leveraging the temporal breadth to capture evolving educational content. The dataset is released under the Open Data Commons Attribution (ODC-By) license, with usage also subject to CommonCrawl's Terms of Use.

\subsection{FineMath}

To enhance mathematical reasoning capabilities, we incorporated FineMath \cite{hugging_face_finemath}, a specialized dataset of high-quality mathematical educational content extracted from CommonCrawl. FineMath is curated using a two-tier quality filtering approach: a regression classifier trained on 1 million annotated web samples assigns educational scores from 0 to 5, evaluating pages based on logical reasoning clarity and step-by-step solution quality. This process results in 34 billion tokens across 21.4 million documents in FineMath-3+, with the option to use the higher-quality FineMath-4+ subset containing 9.6 billion tokens across 6.7 million documents. The dataset underwent additional processing to ensure data quality: documents were deduplicated using MinHash-LSH, filtered to remove non-English content via FineWeb's language classification pipeline, and contamination was addressed by removing samples with 13-gram overlaps against test sets from GSM8k, MATH, MMLU, and ARC, with detailed decontamination logs made publicly available. The dataset is released under the Open Data Commons Attribution License (ODC-By) v1.0, subject to CommonCrawl's Terms of Use.

\subsection{FinePDFs}

To complement web-derived and educational datasets, we incorporated Italian-language content from FinePDFs \cite{kydlicek2025finepdfs}, the largest publicly available corpus built entirely from PDF documents. FinePDFs spans 475 million documents in 1,733 languages, totaling approximately 3 trillion tokens extracted from Common Crawl. PDFs offer distinct advantages over HTML sources by capturing higher-quality, domain-specific content, particularly in legal, academic, and technical domains. The dataset processing pipeline employs a sophisticated two-path extraction strategy: a text-based extraction path using Docling for embedded text, and a GPU-accelerated OCR path using RolmOCR for image-only pages, with an XGBoost-based OCR routing classifier determining the optimal processing path for each document. Language classification was performed using Google/Gemma-3-27b-it on a 20,000-sample subset per language to identify and filter non-Italian content, accounting for code-switching phenomena. Documents were further deduplicated using MinHash-LSH. The dataset undergoes PII anonymization by replacing email addresses and public IP addresses via regex pattern matching; however, as with other web-sourced datasets, some personally identifiable information may remain and removal can be requested via the provided opt-out form. FinePDFs is released under the Open Data Commons Attribution License (ODC-By) v1.0, with usage also subject to CommonCrawl's Terms of Use.

\subsection{StarcoderData}

To enhance code generation capabilities, we incorporated a subset of StarcoderData \cite{bigcode_starcoderdata_2025}, a specialized pretraining dataset consisting of 783 GB of code across 86 programming languages, complemented by 54 GB of GitHub Issues, 13 GB of Jupyter Notebooks, and 32 GB of GitHub commits, totaling approximately 250 billion tokens. Unlike web-derived datasets, StarcoderData is sourced directly from GitHub repositories through The Stack v1.2, a collection of permissively licensed public repositories with copyleft licenses (MPL, EPL, LGPL) excluded and opt-out requests processed through February 9, 2023. The dataset undergoes extensive multi-stage processing: data cleaning combines heuristic filtering and manual inspection, including exclusion of unsupported configuration and programming languages, heuristic filtering of GitHub issues and commits, near-deduplication, and language-specific filtering, with additional exclusion of malicious code. A key differentiator is the comprehensive PII removal approach: in collaboration with Toloka, personally identifiable information including names, passwords, and email addresses is removed from the training data, supplemented by a dedicated PII dataset for training and evaluating PII removal models. The dataset maintains strict license compliance through provenance tracking and provides attribution tools such as integration with VS Code to enable developers to verify generated code against the training data. The Stack is regularly updated to process validated data removal requests, ensuring ongoing respect for contributors' opt-out preferences.

\subsection{Nemotron-Pretraining-SFT-v1}

To further enhance instruction-following and reasoning capabilities, we incorporated Nemotron-Pretraining-SFT-v1 \cite{nvidia2025nvidianemotronnano2}, NVIDIA's synthetic instruction-tuned dataset designed for supervised fine-tuning. Unlike datasets derived from natural web sources, Nemotron-SFT-v1 comprises synthetically generated examples across code, mathematics, general knowledge QA, and fundamental reasoning tasks. STEM data was expanded from high-quality math and science seeds using multi-iteration generation with Qwen3 and DeepSeek models, producing varied and challenging multiple-choice questions with step-by-step solutions, while academic QA pairs were synthesized from complex undergraduate and graduate-level texts. Additional SFT-style data covering code, math, MMLU-style general QA, and fundamental reasoning was generated using DeepSeek-V3 and Qwen3 for logical, analytical, and reading comprehension questions. The dataset is governed by the NVIDIA Data Agreement for Model Training, with the important caveat that any AI model trained with this data that is distributed or made publicly available becomes subject to redistribution and usage requirements under the Qwen License Agreement and DeepSeek License Agreement.

%% file: Sections/12_appendixB.tex
\section{Pretraining Data Filtering}
\label{appendix:pretrain_filtering}

In this section, we describe the filtering procedures and methodologies applied during the pre‑training phase of EngGPT 2, with the objective of ensuring transparency, accountability, and compliance with the EU AI Act. 

Since the detailed composition of data sources and collection methodologies is documented in a separate chapter, the present section focuses solely on:
\begin{itemize}
    \item filtering techniques applied to web‑scale pre‑training corpora
    \item the additional safeguards designed to minimize the ingestion of copyrighted material
    \item the implementation of opt‑out mechanisms for future model training.
\end{itemize} 

These processes constitute a core component of EngGPT 2’s compliance posture, demonstrating that Engineering has undertaken reasonable and demonstrable efforts to use only data that can be lawfully processed for LLM training.

\subsection{Baseline Filtering Pipeline}
EngGPT 2 benefits from an initial preprocessing layer already applied by the dataset providers, which includes standard quality and safety filtering steps such as deduplication, language identification, and boilerplate removal. This baseline provides an initial level of dataset cleaning before the application of copyright-specific safeguards.

\subsection{EngGPT 2 Copyright‑Focused Filtering Pipeline}
To strengthen compliance with the AI Act’s transparency and intellectual‑property obligations, we designed and executed an additional filtering pipeline applied on top of pre-train datasets, targeting specifically copyrighted and editorial content. This pipeline combines rule‑based heuristics, pattern matching, and machine‑learning classification, producing a risk score used to exclude high‑risk records. 

The main components are as follows.
\paragraph{Domain Level Copyright Exclusion List}
A curated domain blacklist was constructed covering digital magazines, news outlets, ebook repositories, publishers and editorial platforms, as well as academic and scientific publishers. 

Any URL belonging to a listed domain is removed during preprocessing; see \autoref{tab:url_list} for a non-exhaustive list of such domains.

\begin{table}[h]
\centering
\setlength{\arrayrulewidth}{0.4pt} 
\setlength{\tabcolsep}{15pt} 
\renewcommand{\arraystretch}{2} 

\begin{tabular}{c|c}
\textbf{Category} & \shortstack{\textbf{Link List}} \\
\specialrule{1.5pt}{0pt}{0pt} 
\shortstack{\textbf{Media \& News}} & European Newspapers Directory – Browse National \& Regional News by Country\tablefootnote{\href{https://newspapers-europe.eu/}{https://newspapers-europe.eu/}}  \\
\textbf{Academic} & sciencedirect.com, nature.com, tandfonline.com \\
\textbf{Broadcasting} & rai.it, bbc.co.uk, zdf.de, rtve.es \\
\textbf{Legal} & wolterskluwer.it, chbeck.de, dalloz.fr \\
\textbf{Literary} & mondadori.it, hachette.fr, penguin.co.uk \\
\textbf{Professional} & ipsoa.it, de jure.it, beck-online.de
\end{tabular}

\vspace{5mm}
\caption{URL List}
\label{tab:url_list}
\end{table}

\paragraph{Editorial Pattern Detection}
Many copyrighted works exhibit identifiable editorial markers. Using lexical rules and NLP heuristics, we exclude records containing chapter markers (e.g., “Chapter 3”, “Volume I”), book‑specific structures (preface, acknowledgements, table of contents), formal article headings used by newspapers or journals, explicit page numbering sequences typical of scanned or reprinted works.

\paragraph{Copyright Notice \& Legal Boilerplate Detection}
We employ a set of regular expressions and pattern‑based detectors to remove records containing explicit copyright symbols (©, ®, ™), “All rights reserved” disclaimers, license statements, publication notices, and legal disclaimers, standard boilerplate text commonly found in copyrighted books or journals.

\paragraph{Composite Copyright Risk Score}
Each record receives a risk score combining some of the previous signals: domain blacklist match, editorial pattern detection, legal boilerplate markers and ISBN/DOI patterns. Records exceeding the risk threshold are removed from the pre‑training corpus.

\subsection{Results}
The copyright-aware pipeline was applied to the full dataset and the process resulted in a substantial reduction of content with high likelihood of being copyrighted, systematic exclusion of entire categories of editorial and scientific content, and improved dataset transparency, auditability, and traceability through a fully reproducible preprocessing procedure.





\begin{table}[h]
\centering
\setlength{\arrayrulewidth}{0.4pt} 
\setlength{\tabcolsep}{15pt} 
\renewcommand{\arraystretch}{2} 

\begin{tabular}{c|c}
\textbf{Metric}  & \textbf{Percentage}\\
\specialrule{1.5pt}{0pt}{0pt} 
\textbf{Clean Records}  & $76.26\%$ \\
\shortstack{\textbf{Flagged records} \\ ($\ge 1$ signal)}  & $ 23.74\% $ \\
\shortstack{\textbf{High-risk records} \\ ($\ge 0.5$ signal)} & $ 1.65\% $   \\
\end{tabular}

\vspace{5mm}
\caption{Copyright Filtering Statistics}
\label{tab:copyright_filtering_statistics}
\end{table}

The most frequent infringement categories were Media \& News ($7.88\%$), Broadcasting ($6.27\%$), ISBN markers ($5.85\%$), and Copyright boilerplate ($5.16\%$), confirming that the pipeline effectively targets editorial and protected sources.

Two alternative risk thresholds were evaluated to balance coverage and precision.
With a threshold of 0.3, 18.8\% of the corpus is excluded, capturing a broader range of potentially copyrighted content.
With a threshold of 0.4, 5.9\% of the corpus is excluded, resulting in a more conservative, higher‑precision filtering strategy.
Both configurations are reported for transparency, enabling downstream policy decisions without altering the underlying technical pipeline.


Overall, the proposed filtering framework represents a key technical contribution of EngGPT 2 toward responsible and compliant large-scale language model training. By integrating baseline quality controls with a dedicated copyright-aware pipeline, EngGPT 2 establishes a transparent, auditable, and reproducible methodology for dataset construction. This approach demonstrates that regulatory requirements under the EU AI Act can be operationalized through concrete engineering solutions, enabling scalable model development while significantly reducing exposure to copyrighted and editorial content. 

\subsection{Opt‑Out Procedures} 
To respect data subject rights and ensure compliance with the AI Act and EU copyright regulations, we implemented an opt-out procedure. If you think that we inadvertently used your copyrighted data, send us an email to enggpt-team@eng.it. We will promptly exclude the relevant data from future model iterations.

%% file: Sections/13_appendixC.tex
\section{Reasoning Template Examples}
\label{app:reasoning_templates}

This appendix reports illustrative examples of the reasoning templates used during training and inference across all supported configurations.

\subsection{Full Reasoning (English)}

\begin{verbatim}
/reasoning_en
<think>
Step-by-step reasoning in English...
</think>
Final answer.
\end{verbatim}

\subsection{Full Reasoning (Italian)}

\begin{verbatim}
/reasoning_ita
<think>
Ragionamento passo per passo in italiano...
</think>
Risposta finale.
\end{verbatim}

\subsection{Turbo Reasoning (English)}

\begin{verbatim}
/reasoning_en
/turbo
<think>
Compressed reasoning trace in English...
</think>
Final answer.
\end{verbatim}

\subsection{Turbo Reasoning (Italian)}

\begin{verbatim}
/reasoning_ita
/turbo
<think>
Ragionamento sintetico in italiano...
</think>
Risposta finale.
\end{verbatim}

\subsection{Non-Reasoning Mode}

When reasoning is disabled (\texttt{enable\_thinking = false}), structural tags are preserved for formatting consistency, but no intermediate reasoning content is generated.

\begin{verbatim}
<think>
</think>
Final answer only.
\end{verbatim}

%% file: Sections/14_appendixD.tex
\section{SFT Data Details}
\label{app:sft_data_details}
This appendix provides a detailed overview of the task taxonomy used to describe the functional coverage of the Training Dataset. The taxonomy is not intended as a normative constraint on dataset construction; rather, it serves as a unifying abstraction that highlights the breadth of behaviors, competencies, and interaction patterns represented across the mixture.
The taxonomy organizes tasks into a structured hierarchy consisting of:

Primary task families, capturing broad functional domains (e.g., reasoning, instruction following, transformation, creative generation).
High‑level categories, representing coherent groups of behaviors within each family.
Sub‑tasks, reflecting concrete interaction types and data patterns observed across the corpus.

This framework enables a systematic interpretation of dataset content and supports clearer comparisons across modalities and data sources.

A detailed breakdown of the taxonomy—including all primary families, categories, and sub‑tasks—is presented in \autoref{tab:task_taxonomy}. This structure is intended to help contextualize how the dataset mixture supports the model’s broad behavioral capabilities and generalization properties.

\begin{table}[h]
\centering
\small
\setlength{\arrayrulewidth}{0.4pt}
\renewcommand{\arraystretch}{1.25}
\setlength{\tabcolsep}{4pt}

\begin{tabular}{p{0.15\linewidth} | p{0.25\linewidth} | p{0.5\linewidth}}

\textbf{Category} & \textbf{Task Type} & \textbf{Sub-Tasks} \\
\noalign{\hrule height 1pt}

Text & Creative Writing &
storytelling; talk about a theme; email writing; trip planner; dialog generation; QA generation; document generation; brainstorming \\

\cline{2-3}

& Text Processing &
rephrase; summarization; translation; NER; text filtering; text masking; text classification; focus shift; style adaptation; text expansion; text projection \\

\cline{2-3}

& Information\newline Extraction &
grounded QA; not-grounded QA; keyword extraction; keyphrase extraction; needleHaystack \\

\cline{2-3}

& Structured Output &
json-like output; markdown output \\

\noalign{\hrule height 0.8pt}

Code & Code Generation &
python; javascript; etc. \\

\cline{2-3}

& Code Migration &
language-to-language migration; framework migration; version upgrade; code refactoring; legacy code modernization; dependency replacement \\

\cline{2-3}

& Code Related Generation &
code documentation; inline comments generation; test case generation; unit/integration tests creation; bug explanation; code review \& suggestions \\

\cline{2-3}

& SQL-like Generation &
query generation; query reformulation; query classification \\

\noalign{\hrule height 0.8pt}

Instruction\newline Following & ActLike &
act-like personas; act-like with style constraints; act-like tool of classification; act-like a console; act-like a tool \\

\noalign{\hrule height 0.8pt}

Alignment & Policies &
safety policy enforcement; content moderation; self-awareness handling; refusal \& self-completion; chit-chat handling; prompt injection detection; jailbreak resistance \\

\cline{2-3}

& Metrics &
KPI-style \\

\noalign{\hrule height 0.8pt}

Data & Data Management &
table QA; time series analysis \\

\noalign{\hrule height 0.8pt}

Problem Solving & Problem Solving &
math; logic; puzzles; science problems; medical questions; etc. \\

\noalign{\hrule height 0.8pt}

Agentic AI & Tool Calling &
tool selection; tool calling; parameter extraction; multi-tool chaining; tool error handling; tool output interpretation \\

\cline{2-3}

& Agentic \newline Orchestration &
intent classification; agent selection; task decomposition; plan generation; multi-agent coordination; reflection \& self-correction; memory usage \& retrieval \\

\noalign{\hrule height 0.8pt}

Misc & Misc &
miscellaneous \\

\end{tabular}

\vspace{5mm}
\caption{Task taxonomy.}
\label{tab:task_taxonomy}
\end{table}

%% file: Sections/15_appendixE.tex
\section{Evaluation Configuration Parameters}
\label{app:eval_config}

The configurations used for the evaluation comparison are reported below.

For a strict cross-model comparison, all models were evaluated using the same generation hyperparameters: temperature = 0.6, top-p = 0.95, top-k = 20, and min-p = 0.

In addition, each model was evaluated in its best-performing configuration to ensure that the reported results reflect the maximum achievable capability under realistic serving conditions. When supported, reasoning mode was enabled during evaluation, and the official serving configurations provided in the model documentation were adopted. References to the corresponding documentation for each best-performing configuration are also provided.

The complete set of serving parameters is reported in \autoref{tab:opt_serving_parameters}.

\begin{table}[h]
    \centering
    \setlength{\arrayrulewidth}{0.4pt}
    \setlength{\tabcolsep}{10pt} 
    \renewcommand{\arraystretch}{1.8}
    
    \begin{tabular}{l|c|c|c|c} 
        & \textbf{Temperature} & \textbf{Top\_p} & \textbf{Top\_k} & \textbf{Min\_p} \\
    \cline{1-5}
    EngGPT2-16B-A3B & 0.6 & 0.95 & 20 & 0 \\
    Qwen3-14B\tablefootnote{\href{https://huggingface.co/Qwen/Qwen3-14B\#best-practices}{https://huggingface.co/Qwen/Qwen3-14B\#best-practices}} & 0.6 & 0.95 & 20 & 0 \\
    Qwen3-30B-A3B\tablefootnote{\href{https://huggingface.co/Qwen/Qwen3-30B-A3B\#best-practices}{https://huggingface.co/Qwen/Qwen3-30B-A3B\#best-practices}} & 0.6 & 0.95 & 20 & 0 \\
    Gpt-oss-20b-high\tablefootnote{\href{https://unsloth.ai/docs/models/gpt-oss-how-to-run-and-fine-tune\#recommended-settings}{https://unsloth.ai/docs/models/gpt-oss-how-to-run-and-fine-tune\#recommended-settings}} & 1 & 1 & 0 & -- \\
    Llama-3.1-8B-Instruct\tablefootnote{\href{https://huggingface.co/meta-llama/Meta-Llama-3-8B-Instruct}{https://huggingface.co/meta-llama/Meta-Llama-3-8B-Instruct}. We evaluated the new model version: Llama-3.1-8B-Instruct \cite{llama3_8b_instruct}.} & 0.6 & 0.9 & -- & -- \\
    gemma-2-9b-it\tablefootnote{\href{https://huggingface.co/google/gemma-2-9b\#advanced-usage}{https://huggingface.co/google/gemma-2-9b\#advanced-usage}} & 1 & -- & -- & -- \\
    gemma-3-12b-it\tablefootnote{\href{https://unsloth.ai/docs/models/gemma-3-how-to-run-and-fine-tune\#recommended-inference-settings}{https://unsloth.ai/docs/models/gemma-3-how-to-run-and-fine-tune\#recommended-inference-settings}} & 1 & 0.95 & 64 & 0 \\
    Moonlight-16B-A3B-Instruct\tablefootnote{No official docs} & 0.6 & 0.95 & 20 & 0 \\
    \end{tabular}
    \vspace{5mm}
    \caption{Optimal serving parameters and reference sources for all evaluated models, as used in the final evaluation. If no value is reported for a given parameter, the default setting from the lm-evaluation-harness framework was used.}
    \label{tab:opt_serving_parameters}
\end{table}